%% file: neurips_2026.tex
\documentclass{article}

\usepackage[eandd, final]{neurips_2026}

\usepackage[utf8]{inputenc}
\usepackage[T1]{fontenc}
\usepackage{hyperref}
\usepackage{url}
\usepackage{booktabs}
\usepackage{amsfonts}
\usepackage{amsmath}
\usepackage{amssymb}
\usepackage{nicefrac}
\usepackage{microtype}
\usepackage{xcolor}
\usepackage{graphicx}
\usepackage{subcaption}
\usepackage{multirow}
\usepackage{makecell}
\usepackage{tablefootnote}
\usepackage{enumitem}
\usepackage{bm}
\usepackage{algorithm}
\usepackage{algpseudocode}
\usepackage{amsthm}

\theoremstyle{plain}

\theoremstyle{definition}

\newcommand{\phoebi}{\textsc{PHOEBI}}
\newcommand{\methodA}{\textsc{SimplexUnmix}}
\newcommand{\methodB}{\textsc{ProtoMatch}}
\newcommand{\methodC}{\textsc{ChannelGroup}}
\newcommand{\R}{\mathbb{R}}

\newcommand{\E}{\mathbb{E}}
\newcommand{\Prob}{\mathbb{P}}
\newcommand{\simplex}{\Delta^{K-1}}
\newcommand{\proto}{\mathbf{P}}
\newcommand{\sparsemax}{\operatorname{sparsemax}}
\newcommand{\ind}{\mathbf{1}}

\newcommand{\SV}[1]{}
\newcommand{\AB}[1]{}

\title{\phoebi{}: An Open-World Benchmark for Multi-Label Bacterial Identification in \\Phase-Contrast Microscopy}

\author{%
  Aaditya Baranwal\quad Md Jahid Hasan\quad Shruti Vyas\\
  Institute of Artificial Intelligence, University of Central Florida, Orlando, FL, USA\\
  \texttt{\{aaditya.baranwal, mdjahid.hasan, shruti\}@ucf.edu}
}

\begin{document}

\maketitle

\input{sections/0_abstract}
\input{sections/1_intro}
\input{sections/2_related}
\input{sections/3_benchmark}
\input{sections/4_experiments}

\input{sections/6_conclusion}

\begin{ack}
This work received no external funding. The authors declare no competing interests.
\end{ack}

\bibliographystyle{plainnat}
\bibliography{references}

\clearpage
\appendix
\input{sections/x_supplementary}


\end{document}

%% file: sections/0_abstract.tex
{\centering
\captionsetup{hypcap=false}
\includegraphics[width=0.85\linewidth]{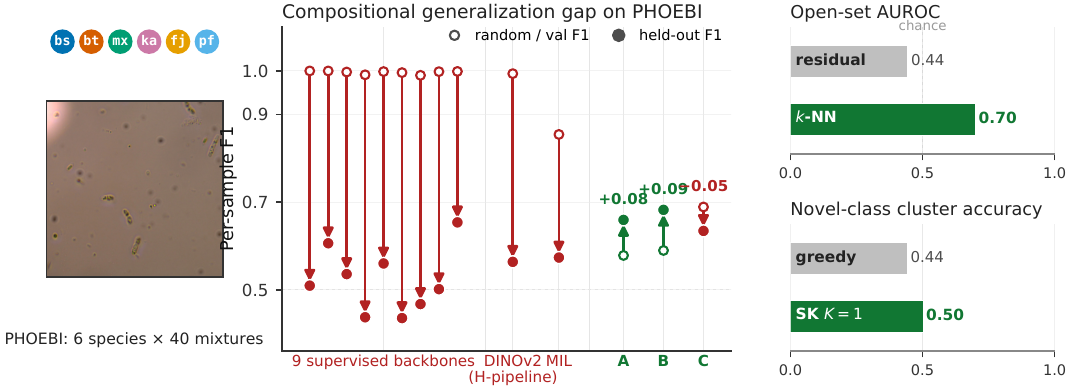}
\captionof{figure}{\textbf{\phoebi{} at a glance.} \textit{Left:} a phase-contrast field from a culture of all six species. \textit{Centre:} each arrow runs from a model's F1 on mixtures seen in training (open circle) to its F1 on mixtures it has never seen (filled circle), so a long downward arrow marks a model that does not carry over to new mixtures. \textit{Right:} the same frozen features, with no further training, flag images that hold an unseen species (top) and gather them into a new class (bottom). Green marks the methods we adopt, and red and grey mark the alternatives they are compared with.}
\label{fig:teaser}
\par}

\begin{abstract}

Optical microscopy lets us watch living bacteria in real time without stains or labels, and it remains the standard instrument for identifying species across clinical, environmental and industrial microbiology. Yet real samples rarely hold a single species. They arrive as mixtures and sometimes carry an organism no model has ever seen. What happens when a model trained on familiar mixtures meets one it has never encountered? No benchmark has asked this of phase-contrast images, so we built one. \phoebi{}\footnote{Code: \href{https://github.com/eternal-f1ame/phoebi}{\nolinkurl{github.com/eternal-f1ame/phoebi}} Data: \href{https://huggingface.co/datasets/sochastic/PHOEBI}{\nolinkurl{huggingface.co/datasets/sochastic/PHOEBI}}} brings together $120{,}000$ phase-contrast images of $40$ mixtures of six bacterial species that we cultured and imaged ourselves, and it holds whole mixtures out of training so that every test is a combination the model has never seen. The answer is sobering. Classifiers that are nearly perfect on the mixtures they know collapse on new ones, and no choice of backbone or pretraining rescues them. The features turn out to be sound and the failure lies in how the evidence within each image is pooled. A culture looks the same everywhere, and decoders built on this simple fact hold steady where large networks fail even though one of them is never trained by gradient descent. The same features go further still. They notice when a species outside the catalogue appears, and the simplest way of giving it a class of its own proves the most effective while leaving the known species almost untouched. \phoebi{} turns the mixed culture, the everyday case at the bench, into a measurable target for open-world recognition.
\end{abstract}

%% file: sections/1_intro.tex
\section{Introduction}
\label{sec:intro}


Bacteria are central to human welfare: they drive bioprocesses such as fermentation and pharmaceutical production, and are tracked as contaminants across food, water, clinical and industrial pipelines. Identifying which species are present is therefore a foundational task across microbiology, and in practice it begins by looking at the sample under a microscope.

\begin{figure}[t]
\centering
\includegraphics[width=\linewidth]{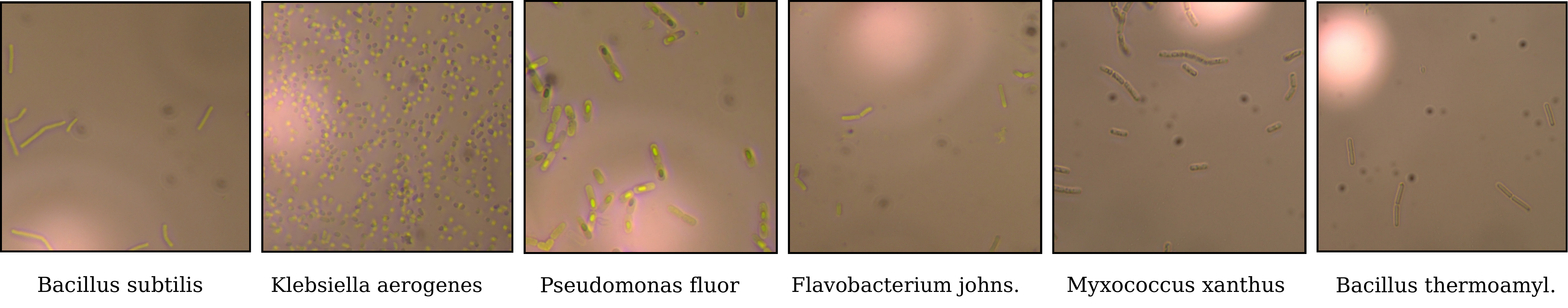}
\caption{Pure cultures of the six \phoebi{} species, with the codes used throughout: \texttt{bs} (\emph{Bacillus subtilis}), \texttt{bt} (\emph{Bacillus thermoamylovorans}), \texttt{mx} (\emph{Myxococcus xanthus}), \texttt{ka} (\emph{Klebsiella aerogenes}), \texttt{fj} (\emph{Flavobacterium johnsoniae}) and \texttt{pf} (\emph{Pseudomonas fluorescens}). The pair to watch is \texttt{bs} and \texttt{bt}, thin rods that overlap in width and density, while the short and encapsulated \texttt{ka} stands apart.}
\label{fig:species-gallery}
\end{figure}

A research swab, a soil isolate, a spoiled-food sample: each arrives at the microbiology bench as a phase-contrast microscopy (PCM) image of a slide-mounted culture, and the operative question is which species are present, and whether one of them might be unseen. Such samples are multi-label by construction because cultures are mixtures, open-world because novel organisms are routine in field-collected material, and fine-grained because closely related rods overlap in shape and differ only in cell-length statistics by factors of two (Figure~\ref{fig:species-gallery}). Visual identification is demanding even for trained microscopists, so automation is the natural path to throughput. Existing bacterial computer-vision (CV) datasets~\citep{zieliski2017deep,treebupachatsakul2019bacteria} are colony-scale, pure-culture or single-label (Table~\ref{tab:dataset-comparison}), so no public benchmark evaluates these three axes together.

Standard bacterial identification today relies on 16S rRNA gene sequencing: sample collection, DNA extraction, PCR amplification, library construction, sequencing, and bioinformatic analysis. The workflow is highly accurate but laboratory-intensive, costly, and delivers results on a timescale of hours to days. PCM offers a faster, label-free alternative that is already routine at the bench.

We introduce the Phase-contrast Optical bEnchmark for Bacterial Identification (\phoebi{}), a benchmark of PCM images of mixed cultures of six rod-shaped species, every culture prepared and imaged in-house (Section~\ref{subsec:dataset}).
Its leave-combinations-out protocol asks the question a working laboratory faces: does a model trained on catalogued mixtures recognise the species in a mixture it has never seen? For the standard fine-grained-recognition pipeline the answer is no. Gradient-trained per-image aggregators, whether fine-tuned end to end, fine-tuned through our own front-end or trained on frozen features, collapse on unseen combinations, and no change of backbone or front-end resolves this \emph{compositional collapse} (Section~\ref{subsec:compgap}). We trace it to the aggregator, which fits the joint label distribution of the combinations it was trained on, and propose three lightweight anchor-based decoders that instead read each species against a fixed geometric prototype over a shared frozen \textsc{DINOv2}~\citep{oquab2023dinov2} tile-feature pool, a design licensed by the spatial homogeneity of slide-mounted cultures (Section~\ref{subsec:setup}). The anchor decoders remain stable under the same protocol, and the same frozen pool supports open-set rejection and novel-class discovery without retraining (Section~\ref{subsec:openworld}).

This work contributes (Figure~\ref{fig:teaser}) \emph{(i)} \phoebi{} with its leave-combinations-out and development protocols (Section~\ref{sec:benchmark}), \emph{(ii)} the compositional collapse of gradient-trained per-image aggregators (Section~\ref{subsec:compgap}), and \emph{(iii)} anchor-based decoders that stay stable under the same shift and extend to the open world on the same frozen features (Sections~\ref{subsec:setup} and~\ref{subsec:openworld}).


%% file: sections/2_related.tex
\section{Related Work}
\label{sec:related}

\input{tables/tab_dataset_comparison}






Bacterial identification from microscopy has been pursued across several imaging modalities. Scanning electron microscopy delivers sub-nanometre surface morphology, but requires chemical fixation and dehydration, precluding live-cell imaging and limiting throughput. Colony-level classification from bright-field or phase-contrast images of agar plates~\citep{bhattacharya2025microcolony} operates on centimetre-scale colonies rather than individual cells and does not compose with liquid-culture identification where discrete colonies are absent. Fluorescence microscopy with species-specific probes enables selective labelling but requires reagents and staining protocols absent from routine workflows. Phase-contrast OM is the natural complement for live-cell work: it is label-free, requires no sample preparation beyond slide mounting, resolves bacterial cells in the $1$--$10\,\mu\mathrm{m}$ size range, and is the standard instrument for monitoring live cultures in bioprocess control, food safety, and environmental surveillance. 
Table~\ref{tab:dataset-comparison} compares \phoebi{} with the existing computer-vision bacterial benchmarks~\citep{zieliski2017deep,treebupachatsakul2019bacteria}, and the most recent work in this vein frames the problem as domain adaptation across optical conditions rather than detection in mixed cultures~\citep{bhattacharya2025microcolony}. The wider cellular-microscopy literature~\citep{moen2019deep,caicedo2017data} likewise assumes single-class images, and recent vision-language microscopy benchmarks~\citep{lozano2024microbench} evaluate generalist vision-language models on closed-set visual question answering over single-label images.

On the computer-vision side \phoebi{} sits at the intersection of multi-label fine-grained recognition, open-set recognition, and novel-class discovery, and adopts established primitives from each. From the multi-label fine-grained literature~\citep{wang2016cnnrnn,liu2021query2label,ridnik2021asl,plantclef2024,fungiclef2024,animalclef2024} we adopt the per-sample F1 read-out popularised by PlantCLEF, but invert the closed-set, search-for-an-object-subregion operating mode. A culture looks the same everywhere, so every crop is a sample of the whole-image label rather than a localization target, and held-out species must be flagged as unknown rather than coerced into a wrong known label. From the open-set literature~\citep{hendrycks2017baseline,liu2020energy,sun2022knn,ruff2021unifying} we adopt the non-parametric $k$-NN cosine-distance tail of~\citet{sun2022knn}. From novel and generalized category discovery~\citep{han2019learning,fini2021unified,vaze2022generalized,li2023modeling,gu2023class,liu2024ufgncd} we adopt the Sinkhorn-Knopp~\citep{cuturi2013sinkhorn} doubly-stochastic assignment of UNO~\citep{fini2021unified} for proposing new classes; the channel-grouped discriminative head is inspired by UFG-NCD's mutual-channel head~\citep{liu2024ufgncd}. The simplex-unmixing decoder extends prototypical networks~\citep{snell2017prototypical} in the spirit of sparse coding~\citep{olshausen1996emergence}, NMF~\citep{lee1999nmf}, and hyperspectral unmixing~\citep{bioucasdias2012hyperspectral}: each tile is a non-negative sparse mixture of class prototypes. What is new is the regime, a real-microscopy multi-label \emph{compositional} split on data the wet-lab community can generate, and a decoder design that exploits the spatial homogeneity this regime provides and natural images lack.

%% file: tables/tab_dataset_comparison.tex
\begin{table}[b]
\centering
\small
\caption{\phoebi{} against existing bacterial microscopy datasets. \emph{Level} separates imaging of individual bacteria from imaging of colonies on agar, \emph{Mag.} is total magnification, and \emph{Comb.} marks a protocol that tests generalisation to unseen mixtures.}
\label{tab:dataset-comparison}
\resizebox{\linewidth}{!}{%
\begin{tabular}{l l l r r c c c c}
\toprule
\textbf{Dataset} & \textbf{Modality} & \textbf{Level} & \textbf{Images} & \textbf{Species} & \textbf{Mag.} & \textbf{Multi-label} & \textbf{Comb.} & \textbf{Public} \\
\midrule

\makecell[l]{DIBaS \\ \citep{zieliski2017deep}}                               & Gram-stain     & Individual & $660$              & $33$ & $1000\times$     & --         & --         & \checkmark \\
\makecell[l]{Bacterial morphology \\ \citep{treebupachatsakul2019bacteria}}    & Gram-stain     & Individual & $800$              & $2$  & $1000\times$    & --         & --         & --         \\
\makecell[l]{Microcolony \\ \citep{bhattacharya2025microcolony}}               & Phase-contrast & Colony     & $630$              & $6$  & $60\times$      & --         & --         & --         \\

\midrule

\textbf{\phoebi{} (ours)} & \textbf{Phase-contrast} & \textbf{Individual} & $\mathbf{{\sim}120\text{K}}$ & $\mathbf{6}$ & $\mathbf{1000\times}$ & $\checkmark$ & $\checkmark$ & $\checkmark$ \\
\bottomrule
\end{tabular}%
}
\end{table}

%% file: sections/3_benchmark.tex
\section{Benchmark}
\label{sec:benchmark}
\label{subsec:dataset}


\begin{figure}[t]
    \centering
    \includegraphics[width=\linewidth]{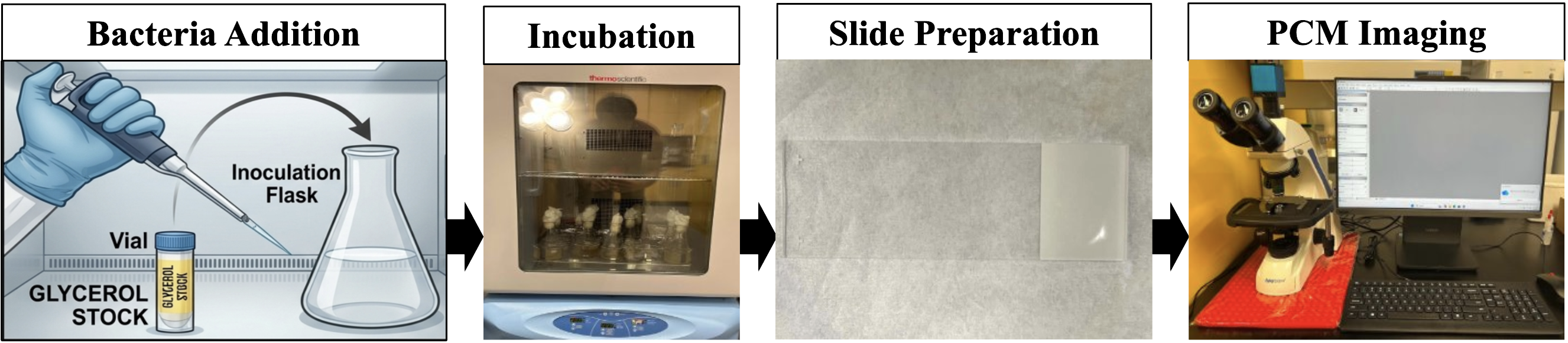}
    \caption{\textbf{Data collection}, from glycerol stock to image: inoculation, incubation, slide preparation and phase-contrast imaging on a single upright microscope.}
    \label{fig:workflow}
\end{figure}

\begin{figure}[t]
    \centering
    \includegraphics[width=\linewidth]{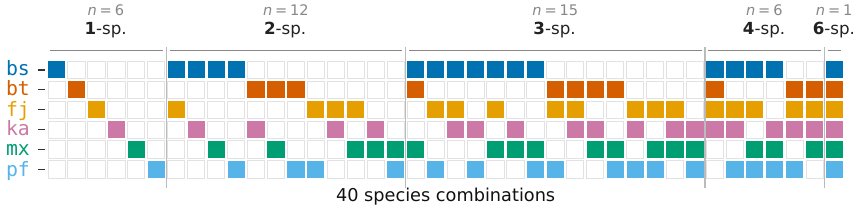}
    \caption{The combinations of \phoebi{}. Each column is one combination and each filled cell a species present in it, with columns grouped by the number of species they mix.}
    \label{fig:dataset-matrix}
\end{figure}

\input{tables/tab_datasets}

The benchmark consists of approximately $120{,}000$ PCM images at $1000\times$ total magnification ($100\times$ oil immersion, $\mathrm{NA}=1.25$) of $40$ cultures that we grew in a sterile laboratory and imaged ourselves, which gives complete control over species composition (Figure~\ref{fig:workflow}). They span the six species of Figure~\ref{fig:species-gallery}. All six are rod-shaped and motile, but they sample three distinct motility mechanisms (peritrichous flagella, polar flagella, gliding), span Gram-positive and Gram-negative, and have cell lengths from $1\,\mu\mathrm{m}$ to $10\,\mu\mathrm{m}$, so the inter-class geometry exercises both easy and morphologically confusable discriminations. The colour camera gives every image the same warm cast, a uniform shift that carries no species cue (Appendix~\ref{app:rgb-cast}). The combinations run from the six single species to the full six-species mixture (Figure~\ref{fig:dataset-matrix}, with per-order counts, split sizes and per-species statistics in Table~\ref{tab:dataset}). Cultures were inoculated from glycerol stocks, grown to characteristic stage and imaged on the same upright phase-contrast microscope (dataset card in Appendix~\ref{app:dataset}). The species are never co-cultured. Each combination is mixed at acquisition time from separately grown and individually verified pure suspensions, so every label is verified at culture level. Beyond that, we bound the rate at which a species misses an individual field by sampling at under $5\%$ on average (Appendix~\ref{app:label-noise}).

\label{subsec:protocol}
We release two evaluation protocols. The in-distribution \emph{random split} divides each combination's frames 80/10/10 in acquisition order, so adjacent frames never straddle a boundary. The leave-combinations-out (LCO) split, around which the experiments are organized, holds out nine entire combinations that range from a single species to the full mixture (\texttt{bt}, \texttt{bs\_pf}, \texttt{ka\_fj}, \texttt{bs\_mx\_fj}, \texttt{bs\_ka\_pf}, \texttt{mx\_ka\_fj}, \texttt{bs\_bt\_ka\_fj}, \texttt{bs\_mx\_fj\_pf} and \texttt{bs\_bt\_mx\_ka\_fj\_pf}). Three constraints shape it. Held-out combinations stay out of training and validation. Every species appears in at least one trained-on combination, so the protocol tests compositional generalization rather than novel-class detection. And the held-out set spans combination orders, so performance can be reported as a function of compositional complexity. The protocol generalizes to any multi-label benchmark with combinatorial label structure. All experiments use the tile front-end described in Section~\ref{subsec:approach} with thresholds calibrated on the in-distribution validation split.

Two further design choices make the protocol's numbers trustworthy. Every image of a combination shares one label vector, so a held-out metric compares combinations rather than images, and we bootstrap intervals over combinations accordingly. And because the validation split is in distribution and the held-out combinations are reserved for reporting, the release includes a development protocol of five nested folds over the trained-on combinations for model selection (Appendix~\ref{app:devprotocol}).

%% file: tables/tab_datasets.tex
\begin{table}[b]
\centering
\small
\caption{The \phoebi{} dataset by combination order and by species.}
\label{tab:dataset}

\begin{minipage}[t]{0.40\linewidth}
\centering
\textbf{(a) Per-order combinations and split sizes.}\\[2pt]
\resizebox{\linewidth}{!}{%
\begin{tabular}{lrrrrr}
\toprule
\textbf{Order} & \textbf{Combos} & \textbf{Images} & \textbf{Train} & \textbf{Val} & \textbf{Test} \\
\midrule
$1$ (single)        &  6 & 18{,}000 & 14{,}400 & 1{,}800 & 1{,}800 \\
$2$ (pair)          & 12 & 36{,}000 & 28{,}800 & 3{,}600 & 3{,}600 \\
$3$ (triple)        & 15 & 45{,}000 & 36{,}000 & 4{,}500 & 4{,}500 \\
$4$ (quadruple)     &  6 & 18{,}000 & 14{,}400 & 1{,}800 & 1{,}800 \\
$6$ (six-species)   &  1 &  3{,}000 &  2{,}400 &    300 &    300 \\
\midrule
\textbf{Total}      & \textbf{40} & \textbf{120{,}000} & \textbf{96{,}000} & \textbf{12{,}000} & \textbf{12{,}000} \\
\bottomrule
\end{tabular}%
}
\end{minipage}%
\hfill
\begin{minipage}[t]{0.58\linewidth}
\centering
\textbf{(b) Per-species morphology and coverage.}\\[2pt]
\resizebox{\linewidth}{!}{%
\begin{tabular}{llllr}
\toprule
\textbf{Tok} & \textbf{Species} & \textbf{Length} & \textbf{Morphology} & \textbf{Images} \\
\midrule
\texttt{bs} & \emph{Bacillus subtilis}        & $4$--$10\,\mu\mathrm{m}$   & slender straight rod                & 51{,}000 \\
\texttt{bt} & \emph{Bacillus thermoamylovorans} & ${\sim}4\,\mu\mathrm{m}$   & slender straight rod                & 39{,}000 \\
\texttt{fj} & \emph{Flavobacterium johnsoniae}  & $5$--$10\,\mu\mathrm{m}$   & mid-length rod, tapered ends        & 57{,}000 \\
\texttt{ka} & \emph{Klebsiella aerogenes}       & $1$--$3\,\mu\mathrm{m}$    & encapsulated short stocky rod       & 57{,}000 \\
\texttt{mx} & \emph{Myxococcus xanthus}         & $5$--$10\,\mu\mathrm{m}$   & mid-length rod, blunt ends          & 57{,}000 \\
\texttt{pf} & \emph{Pseudomonas fluorescens}    & $1.5$--$3\,\mu\mathrm{m}$  & short, straight to slightly curved  & 54{,}000 \\
\bottomrule
\end{tabular}%
}
\end{minipage}

\end{table}

%% file: sections/4_experiments.tex
\section{Anchor-Based Decoders}
\label{subsec:setup}
\label{subsec:approach}

\begin{figure}[t]
    \centering
    \includegraphics[width=\linewidth]{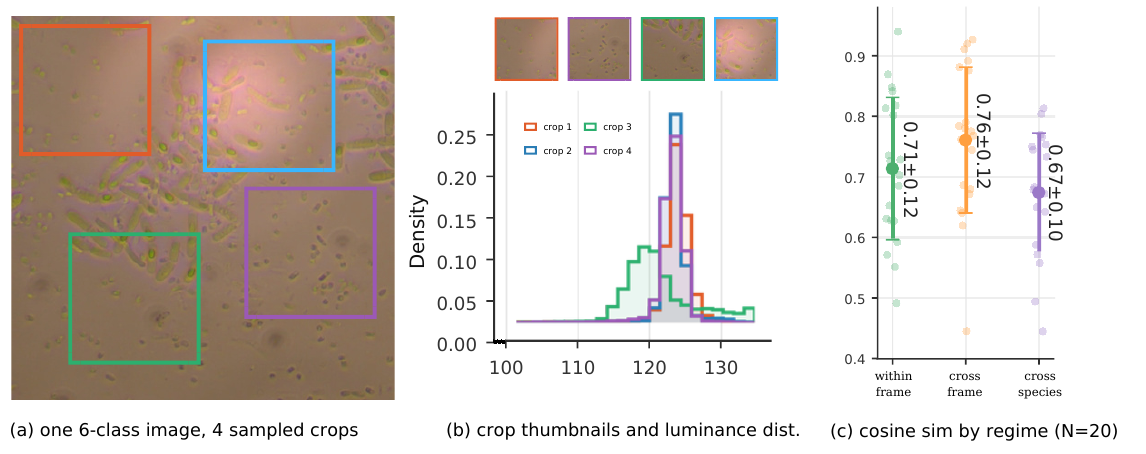}
    \caption{Evidence for Assumption H. The crop colours in (a) match the thumbnails and histograms in (b). In (c) each faint point is one pair of crops, and the solid marker with bars gives the mean and standard deviation over the twenty pairs of each regime.}
    \label{fig:spatial-homogeneity}
\end{figure}

A slide-mounted culture in suspension shows bacterial cells from all present species uniformly dispersed across the field of view (away from the edges), producing a homogeneous view.

This is the feature the framework leans on: there is nothing to localise, and any sufficiently large crop carries the whole-image label. We write $z(c) = \phi(c) / \lVert \phi(c) \rVert_2 \in \mathcal{S}^{D-1}$ for the L$_2$-normalized embedding of a $224 \times 224$ crop $c$ under a frozen feature extractor $\phi$ (\textsc{DINOv2} ViT-S/14, $D = 384$). The framework operates entirely on these tile embeddings, and the backbone is never fine-tuned.

\begin{center}
\fbox{\parbox{0.93\linewidth}{%
\emph{Assumption H (Spatial Homogeneity).} For any image $x$ and any crop $c \subset x$ whose linear size exceeds the longest cell length and whose area contains sufficiently many cells to be representative,
\[
\Prob \bigl( y \mid c \bigr) \;=\; \Prob \bigl( y \mid x \bigr) \;=\; y(x).
\]%
}}
\end{center}

The rest of the framework follows from two consequences of this single property. Random crops become label-preserving augmentations, so training a tile-level classifier with the image label is Bayes-consistent with training an image-level classifier. And per-tile scores average to image-level scores with variance $\mathcal{O}(1/T)$, a rate we verify in Section~\ref{subsec:supporting}. Empirically, the cosine similarity between crops of one image matches that between crops of different images of the same species, and both exceed the similarity across species, so the crops of one image are statistically interchangeable with crops of any image of the same species (Figure~\ref{fig:spatial-homogeneity}, Appendix~\ref{app:math}).

PCM images carry a slowly-varying multiplicative hotspot from the K\"{o}hler-illuminated condenser, which we remove with a per-channel Gaussian background estimate ($\sigma = 64$ px, large enough to capture the lamp gradient while preserving cellular structure). We then sample $T = 16$ tiles of side $s = 224$ per image (uniform random crops at training, a deterministic $4{\times}4$ grid at inference), and embed and normalise each as above. The training recipes and equations are in Appendix~\ref{app:decoder-details}.
The three decoders share identical tile features and differ only in how strongly they commit to the prototype geometry, which lets the experiments isolate what the anchor contributes.

\textbf{\methodA{} (simplex unmixing)} makes the strongest geometric commitment: scaled cosine logits are projected onto the probability simplex via sparsemax~\citep{martins2016sparsemax}, producing exact zeros for absent species. Image-level presence is the mean sparsemax weight across tiles, and the per-tile reconstruction residual is one of the open-set scores of Section~\ref{subsec:openworld}.

\textbf{\methodB{} (cosine matching)} drops the simplex constraint and reads each tile as $K$ independent cosine similarities to the prototypes. Image-level scores are the per-tile mean per class, and per-class thresholds are the $5$th percentile of each class's score over positive validation images. The only learned object is the prototype matrix, and there is no gradient training.

\textbf{\methodC{} (channel-grouped discriminative head)} splits the embedding into $K$ contiguous groups of $64$ channels and trains $K$ linear binary classifiers ($390$ parameters total) with binary cross-entropy and $50\%$ per-species channel dropout~\citep{liu2024ufgncd}. Image-level logits are mean-aggregated and thresholds are calibrated on val by argmax-F1.

\methodA{} and \methodB{} start from the same prototypes, the mean embedding of each species' pure cultures, which already sit near the converged reconstruction loss. \methodA{} trains for thirty epochs from there, and \methodB{} keeps them as they are.

\setcounter{topnumber}{3}\setcounter{totalnumber}{4}
\section{Experiments}
\label{sec:analysis}
\label{subsec:compgap}
\label{subsec:supporting}

\input{tables/tab_supervised_baselines}

LCO emulates the deployment scenario directly: a model trained on the combinations a laboratory has already cultured must keep working when a new combination arrives. We evaluate three families of supervised methods under it (Table~\ref{tab:supervised-baselines}). Nine modern backbones are fine-tuned end to end with a single $\texttt{nn.Linear}(D, K)$ binary cross-entropy (BCE) head, the same \textsc{DINOv2} is fine-tuned through our tile-and-illumination front-end as a pipeline-matched control, and a frozen-feature attention-MIL head~\citep{ilse2018attention} reads the very tile features the \phoebi{} decoders use. All three collapse. The fine-tunes saturate at $0.97$ to $1.00$ F1 in distribution and fall to $0.44$ to $0.61$ on held-out combinations, the pipeline-matched control reproduces the drop, and the MIL head loses over a third of its F1 without ever touching its backbone. The collapse is a property of the aggregator, not the backbone.

\input{tables/tab_encoder_probe}

The choice of encoder matters little. Thirteen frozen encoders, general-purpose and biomedical, score within a narrow band under a multi-label linear probe on the same tiles, and the small \textsc{DINOv2} behind all three decoders sits within $2.3$ points of the best, the $51{\times}$ larger histopathology model Prov-GigaPath (Table~\ref{tab:encoder-probe}, Appendix~\ref{app:encoder-probe}). The signal \phoebi{} measures lives in the imaging modality and the front-end rather than in a pretraining objective.

The collapse also replicates on a four-class subset cultured in a separate batch and imaged in a separate session, where the nine-backbone sweep again saturates in distribution and collapses on held-out combinations by a comparable margin (Table~\ref{tab:supervised-baselines-4class}, Appendix~\ref{app:4class}).

\input{tables/tab_supervised_baselines_4class}

\input{tables/tab_bids_heldout}

\begin{figure}[t]
    \begin{minipage}[t]{0.54\textwidth}
        \centering
        \includegraphics[width=\linewidth]{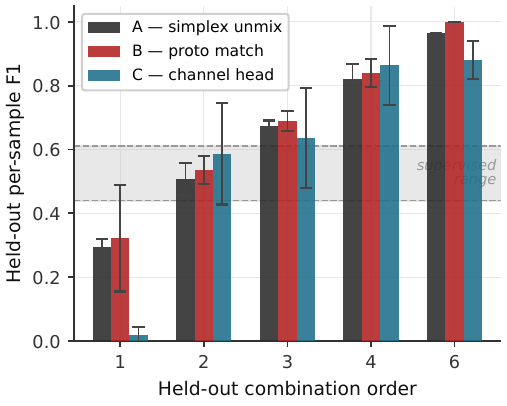}
    \end{minipage}\hfill
    \begin{minipage}[t]{0.40\textwidth}
        \centering
        \includegraphics[width=\linewidth]{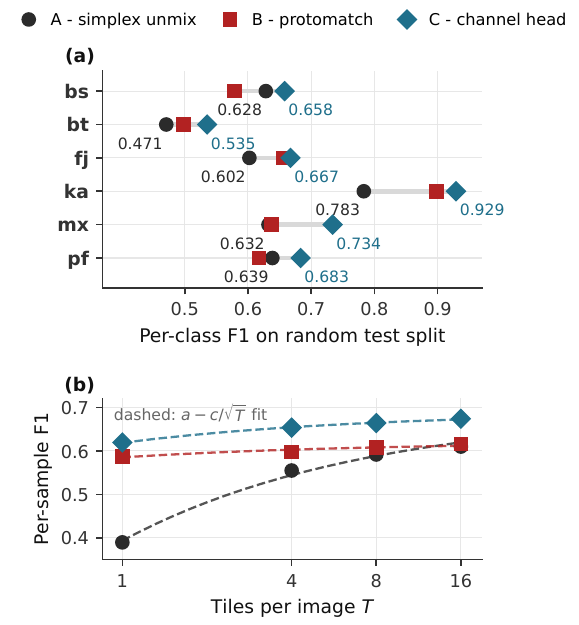}
    \end{minipage}
    \caption{Decoder results. \textit{Left:} held-out per-sample F1 against the number of species in a combination, with error bars over three seeds. The grey band spans the supervised baselines. A constant all-present predictor scores $2k/(k+6)$ at order $k$, so higher orders are easier by construction. \textit{Right:} per-class F1 \textit{(a)} and per-sample F1 against the number of tiles averaged per image \textit{(b)}, both measured in distribution on the random test split, as in Table~\ref{tab:headline}.}
    \label{fig:per-order-f1}
    \label{fig:per-class-f1}
    \label{fig:tile-curve}
\end{figure}

On the six-class LCO split, none of the three \phoebi{} decoders collapses (Table~\ref{tab:bids-heldout}). Held-out combinations contain more species on average than trained ones, which by itself raises per-sample F1, so Tables~\ref{tab:supervised-baselines} and~\ref{tab:bids-heldout} include a constant all-present predictor as the reference level that order composition sets (Figure~\ref{fig:per-order-f1}, left). The gradient-trained aggregators fall below that reference, while the anchor decoders stay at or near it and \methodB{}'s margin above it holds at the level of whole combinations, pooled over all three LCO partitions (Appendix~\ref{app:devprotocol}).

A stronger encoder does not buy compositional generalisation. Re-running the LCO protocol with Prov-GigaPath raises validation F1 but not held-out F1, and widens \methodC{}'s gap (Table~\ref{tab:gigapath-lco}).

\input{tables/tab_gigapath_lco}

The mechanism is structural. A gradient-trained aggregator fits $P(\mathbf{y} \mid \mathrm{combination})$ on the combinations it saw, while an anchor-based decoder reads each species independently against a fixed geometric object and has no per-combination decision surface to overfit. The init-only block of Table~\ref{tab:bids-heldout} isolates this at the decoder level: training the unanchored \methodC{} for thirty epochs lifts validation F1 and \emph{lowers} held-out F1, per-combination overfitting in miniature, while training the simplex-constrained \methodA{} leaves held-out F1 unchanged (Appendix~\ref{app:per-order}).

\input{tables/tab_headline}

In distribution the ordering inverts (Table~\ref{tab:headline}): \methodC{}, the least constrained decoder, leads every column, the trade-off the three decoders were built to expose. Per-class F1 ranks the species consistently across decoders, \texttt{ka} easiest and \texttt{bt} hardest, and per-sample F1 rises monotonically and saturates with the number of tiles, following the $a - c/\sqrt{T}$ curve that $\mathcal{O}(1/T)$ variance predicts under Assumption~H (Figure~\ref{fig:per-class-f1}, right). The assumption also holds at the edge of the field, since averaging only the four interior tiles lowers held-out F1 by no more than the smaller tile count predicts (Appendix~\ref{app:ablations}). Predictions are also invariant to the six axis-aligned rotations and flips, which change \methodB{}'s per-sample F1 by under half a point.

Across acquisition conditions, the decoders are insensitive to the illumination and detector variables that differ most between microscopes. Condenser misalignment, vignetting, a change of lamp intensity or a detector gamma curve costs at most about two points of per-sample F1 at the worst severity (Appendix~\ref{app:acquisition}). The illumination correction buys this invariance for a fraction of a point in distribution, and without it vignetting alone costs \methodB{} $13.6$ points (Table~\ref{tab:illum-matched}). Tiling is just as essential, since a single resized tile in place of the grid costs $21$ to $30$ points. Under focus drift, sensor noise and JPEG re-encoding, which illumination correction does not address, \methodA{} degrades least of the three decoders: its simplex projection discards small perturbations of the tile embedding that a raw cosine read-out passes straight through.

\input{tables/tab_illum_matched}

\label{subsec:openworld}
Beyond presence, the same frozen features support the two open-world tasks a laboratory meets next. Open-set rejection flags an image that contains a species outside the catalogue, and novel-class discovery turns that species into a class of its own. Both run under a leave-one-out cross-validation (LOOCV) protocol over species, which fits \methodA{} on the remaining species from pure cultures and scores the test split against it using only known-class data, with no injected unknowns whose provenance might confound the signal~\citep{hendrycks2019anomaly} (Appendix~\ref{app:loocv-algorithm}).

\begin{figure}[t]
    \centering
    \includegraphics[width=\linewidth]{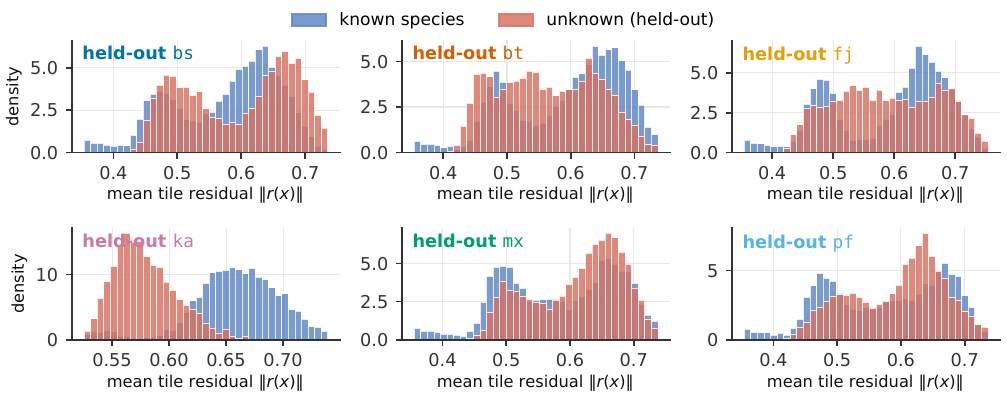}
    \caption{Reconstruction residuals of \methodA{} in each LOOCV fold, for images of known species (blue) and of the held-out species (red). Wherever the two overlap, a residual threshold cannot flag the unknown. Held-out \texttt{ka} even reconstructs better than the known species, because its features lie inside the convex hull of the other five prototypes.}
    \label{fig:residual-hist}
\end{figure}

Open-set rejection shows where the evidence of an unknown species lives. Global functions of the per-class similarity vector, whether residual norm, negative maximum similarity or energy, do not separate held-out species from known ones (Table~\ref{tab:osr-sweep}, Appendix~\ref{app:osr-sweep}), because held-out tiles lie on the same manifold as the retained species (Figure~\ref{fig:residual-hist}). They do occupy distinct local neighbourhoods, which a non-parametric $k$-NN cosine distance to the training tiles~\citep{sun2022knn} recovers, lifting AUROC from chance to $0.70$ and beating the residual on all six folds (Table~\ref{tab:loocv}). Averaging that distance over all tiles works best among the variants selected on an inner validation loop, because an unknown species leaves its trace across the whole image rather than in a few tiles (Appendix~\ref{app:negative}).

Discovery goes one step further and proposes the new class itself. The normalized test tile features are clustered into new prototypes appended to $\proto$, and we measure cluster accuracy and the change in known-class F1. The clustering primitive is decisive (Table~\ref{tab:discovery}). Greedy cosine clustering over-fragments the test pool into dozens of prototypes per fold and degrades known-class F1 substantially. Sinkhorn-Knopp~\citep{cuturi2013sinkhorn} doubly-stochastic assignment at $K = 1$, the natural setting when exactly one species is held out, pools the test tiles into a single prototype. Half of the images that contain the novel species prefer it to every known prototype, and known-class F1 moves an order of magnitude less than under greedy clustering. The simplest primitive is also the strongest, since the more elaborate gradient-based discovery heads of UNO~\citep{fini2021unified} and SimGCD~\citep{wen2023simgcd} and class-relation distillation~\citep{gu2023class} all score below it (Table~\ref{tab:discovery}, Appendix~\ref{app:ncd-negative}).

\input{tables/tab_discovery}

%% file: tables/tab_supervised_baselines.tex
\begin{table}[t]
\centering
\small
\caption{Supervised baselines on the random split and under LCO, where the drop from in-distribution to held-out F1 ($\Delta$F1) measures the collapse. The first row is the constant predictor that marks every species present, and bold marks the best trained model per column. $^{\dag}$Florence-2 DaViT-B never fits the LCO training split and reproduces the constant predictor, so it is left out of the collapse range.}
\label{tab:supervised-baselines}
\resizebox{\linewidth}{!}{%
\begin{tabular}{lc | ccc | cccc | c}
\toprule
 & & \multicolumn{3}{c|}{\textbf{Random $80/10/10$ test}} & \multicolumn{4}{c|}{\textbf{Held-out combinations test}} & \\
\textbf{Backbone} & \textbf{Params (M)} & F1 $\uparrow$ & macro F1 $\uparrow$ & EM $\uparrow$ & in-dist F1 $\uparrow$ & F1 $\uparrow$ & macro F1 $\uparrow$ & EM $\uparrow$ & $\Delta$F1 $\downarrow$ \\
\midrule
All-present predictor (constant) & $0$ & $0.588$ & $0.607$ & $0.025$ & $0.572$ & $0.654$ & $0.674$ & $0.111$ & $-0.083$ \\
\midrule
ResNet-50 & 25.6 & $\mathbf{1.000}$ & $\mathbf{1.000}$ & $\mathbf{0.999}$ & $\mathbf{1.000}$ & $0.509$ & $0.551$ & $0.003$ & $0.491$ \\
ConvNeXt-B & 88.6 & $\mathbf{1.000}$ & $\mathbf{1.000}$ & $\mathbf{0.999}$ & $\mathbf{1.000}$ & $\mathbf{0.606}$ & $\mathbf{0.648}$ & $\mathbf{0.023}$ & $\mathbf{0.394}$ \\
ViT-B/16 IN21k & 86.6 & $0.997$ & $0.998$ & $0.993$ & $0.999$ & $0.536$ & $0.557$ & $0.005$ & $0.463$ \\
DINOv2 ViT-S/14 & 22.1 & $0.991$ & $0.990$ & $0.962$ & $0.989$ & $0.437$ & $0.487$ & $0.000$ & $0.552$ \\
DINOv3 ViT-S/16 & 21.6 & $0.998$ & $0.999$ & $0.996$ & $\mathbf{1.000}$ & $0.560$ & $0.613$ & $0.001$ & $0.440$ \\
CLIP ViT-B/16 & 86.6 & $0.996$ & $0.996$ & $0.984$ & $0.999$ & $0.435$ & $0.464$ & $0.000$ & $0.564$ \\
SigLIP ViT-B/16 & 92.9 & $0.990$ & $0.991$ & $0.970$ & $0.997$ & $0.467$ & $0.528$ & $0.000$ & $0.529$ \\
EVA-02 CLIP B/16 & 86.3 & $0.998$ & $0.998$ & $0.996$ & $0.999$ & $0.501$ & $0.554$ & $0.004$ & $0.498$ \\
Florence-2 DaViT-B$^{\dag}$ & 90.4 & $0.999$ & $0.999$ & $0.997$ & $0.571^{\dag}$ & $0.654$ & $0.674$ & $0.111$ & $-0.082$ \\
\midrule
\multicolumn{10}{l}{\emph{End-to-end fine-tune through the \textup{\phoebi{}} tile + illumination pipeline (control):}} \\
DINOv2 ViT-S/14 (\phoebi{}) & 22.1 & $0.994$ & $0.994$ & $0.975$ & $1.000$ & $0.564$ & $0.619$ & $0.000$ & $0.436$ \\
\midrule
\multicolumn{10}{l}{\emph{Frozen \textup{\textsc{DINOv2}} features with gradient-trained per-image aggregation:}} \\
Attention MIL~\citep{ilse2018attention} & $0.07$ & $0.855$ & $0.874$ & $0.519$ & $0.906$ & $0.574$ & $0.628$ & $0.060$ & $0.332$ \\
\bottomrule
\end{tabular}%
}
\end{table}

%% file: tables/tab_encoder_probe.tex
\begin{table}[t]
\centering
\small
\caption{Thirteen frozen encoders behind the same tile front-end and linear head, on the random test split. General-purpose encoders come first and biomedical ones last, and the six right-hand columns give per-species F1. Bold marks the best per column.}
\label{tab:encoder-probe}
\resizebox{\linewidth}{!}{%
\begin{tabular}{lcc ccc cccccc}
\toprule
\textbf{Encoder} & \textbf{Params (M)} & \textbf{$D$} & \textbf{Per-sample F1} $\uparrow$ & \textbf{Macro F1} $\uparrow$ & \textbf{Exact} $\uparrow$ & \textbf{bs} & \textbf{bt} & \textbf{fj} & \textbf{ka} & \textbf{mx} & \textbf{pf} \\
\midrule
ResNet-50 & 25.6 & 2048 & $0.672$ & $0.698$ & $0.078$ & $0.655$ & $0.529$ & $0.673$ & $0.923$ & $0.713$ & $0.694$ \\
ConvNeXt-B & 88.6 & 1024 & $0.682$ & $0.711$ & $0.088$ & $0.685$ & $0.551$ & $0.685$ & $0.920$ & $0.726$ & $0.698$ \\
ViT-B/16 IN21k & 86.6 & 768 & $0.692$ & $0.718$ & $0.102$ & $0.679$ & $0.557$ & $0.696$ & $0.917$ & $0.742$ & $\mathbf{0.715}$ \\
DINOv2 ViT-S/14 & 22.1 & 384 & $0.676$ & $0.708$ & $0.063$ & $0.675$ & $0.540$ & $0.661$ & $0.928$ & $0.746$ & $0.696$ \\
DINOv3 ViT-S/16 & 21.6 & 384 & $0.671$ & $0.700$ & $0.076$ & $0.656$ & $0.534$ & $0.654$ & $0.921$ & $0.734$ & $0.701$ \\
CLIP ViT-B/16 & 86.6 & 768 & $0.677$ & $0.710$ & $0.071$ & $0.667$ & $0.532$ & $0.690$ & $0.922$ & $0.751$ & $0.701$ \\
SigLIP ViT-B/16 & 92.9 & 768 & $0.666$ & $0.693$ & $0.062$ & $0.648$ & $0.528$ & $0.672$ & $0.899$ & $0.725$ & $0.684$ \\
EVA-02 CLIP B/16 & 86.3 & 768 & $0.663$ & $0.692$ & $0.073$ & $0.648$ & $0.542$ & $0.685$ & $0.906$ & $0.681$ & $0.687$ \\
Florence-2 DaViT-B & 90.4 & 1024 & $0.639$ & $0.665$ & $0.039$ & $0.623$ & $0.511$ & $0.666$ & $0.858$ & $0.660$ & $0.672$ \\
Prov-GigaPath ViT-G & 1135.0 & 1536 & $\mathbf{0.699}$ & $\mathbf{0.720}$ & $\mathbf{0.159}$ & $0.688$ & $\mathbf{0.589}$ & $0.693$ & $\mathbf{0.931}$ & $0.731$ & $0.686$ \\
UNI ViT-L & 303.3 & 1024 & $0.691$ & $0.718$ & $0.102$ & $0.681$ & $0.559$ & $\mathbf{0.702}$ & $0.926$ & $\mathbf{0.770}$ & $0.670$ \\
Phikon ViT-B/16 & 86.4 & 768 & $0.696$ & $0.718$ & $0.109$ & $\mathbf{0.689}$ & $0.579$ & $0.696$ & $0.921$ & $0.737$ & $0.690$ \\
BiomedCLIP ViT-B/16 & 195.9 & 512 & $0.685$ & $0.711$ & $0.050$ & $0.680$ & $0.577$ & $0.668$ & $0.922$ & $0.727$ & $0.694$ \\
\bottomrule
\end{tabular}%
}
\end{table}

%% file: tables/tab_supervised_baselines_4class.tex
\begin{table}[t]
\centering
\small
\caption{The sweep of Table~\ref{tab:supervised-baselines} repeated on four of the species (\texttt{bs}, \texttt{fj}, \texttt{ka} and \texttt{pf}), cultured in a separate batch and imaged in a separate session. Bold marks the best trained model per column.}
\label{tab:supervised-baselines-4class}
\resizebox{\linewidth}{!}{%
\begin{tabular}{lc | ccc | cccc | c}
\toprule
 & & \multicolumn{3}{c|}{\textbf{Random $80/10/10$ test}} & \multicolumn{4}{c|}{\textbf{Held-out combinations test}} & \\
\textbf{Backbone} & \textbf{Params (M)} & F1 $\uparrow$ & macro F1 $\uparrow$ & EM $\uparrow$ & in-dist F1 $\uparrow$ & F1 $\uparrow$ & macro F1 $\uparrow$ & EM $\uparrow$ & $\Delta$F1 $\downarrow$ \\
\midrule
ResNet-50 & 25.6 & $\mathbf{1.000}$ & $\mathbf{0.999}$ & $0.997$ & $\mathbf{1.000}$ & $0.642$ & $0.672$ & $0.050$ & $0.357$ \\
ConvNeXt-B & 88.6 & $0.999$ & $0.999$ & $\mathbf{0.998}$ & $\mathbf{1.000}$ & $0.673$ & $\mathbf{0.696}$ & $0.026$ & $0.326$ \\
ViT-B/16 IN21k & 86.6 & $0.998$ & $0.998$ & $0.993$ & $0.998$ & $\mathbf{0.680}$ & $0.673$ & $\mathbf{0.145}$ & $\mathbf{0.318}$ \\
DINOv2 ViT-S/14 & 22.1 & $0.996$ & $0.996$ & $0.987$ & $0.997$ & $0.633$ & $0.660$ & $0.046$ & $0.362$ \\
DINOv3 ViT-S/16 & 21.6 & $0.999$ & $0.999$ & $0.994$ & $\mathbf{1.000}$ & $0.505$ & $0.548$ & $0.001$ & $0.493$ \\
CLIP ViT-B/16 & 86.6 & $0.997$ & $0.998$ & $0.992$ & $0.998$ & $0.663$ & $0.671$ & $0.011$ & $0.334$ \\
SigLIP ViT-B/16 & 92.9 & $0.998$ & $0.997$ & $0.992$ & $1.000$ & $0.653$ & $0.653$ & $0.117$ & $0.345$ \\
EVA-02 CLIP B/16 & 86.3 & $0.998$ & $0.998$ & $0.992$ & $1.000$ & $0.628$ & $0.629$ & $0.011$ & $0.369$ \\
Florence-2 DaViT-B & 90.4 & $0.999$ & $0.999$ & $\mathbf{0.998}$ & $0.996$ & $0.573$ & $0.639$ & $0.027$ & $0.427$ \\
\bottomrule
\end{tabular}%
}
\end{table}

%% file: tables/tab_bids_heldout.tex
\begin{table}[t]
\centering
\small
\caption{\phoebi{} decoders under the LCO protocol of Table~\ref{tab:supervised-baselines}. Trained rows are mean $\pm$ std over three seeds, and init-only rows use the closed-form prototypes with no training. $\Delta$F1 is validation minus held-out F1. Held-out combinations hold more species on average, so even the constant predictor gains on them, and each decoder is best read against that row. Combination-level intervals are in Appendix~\ref{app:devprotocol}. Bold marks the best trained decoder per column.}
\label{tab:bids-heldout}
\resizebox{\linewidth}{!}{%
\begin{tabular}{lcccc}
\toprule
\textbf{Decoder} & \textbf{In-dist Val F1} $\uparrow$ & \textbf{Held-out F1} $\uparrow$ & \textbf{Held-out Macro F1} $\uparrow$ & \textbf{$\Delta$F1} $\downarrow$ \\
\midrule
All-present predictor (constant) & $0.572$ & $0.654$ & $0.674$ & $-0.083$ \\
\midrule
Image-level e2e (Tab.~\ref{tab:supervised-baselines}, range) & $0.97$--$1.00$ & $0.44$--$0.61$ & $0.46$--$0.65$ & $0.39$ to $0.57$ \\
DINOv2 e2e via \phoebi{} pipeline & $1.000$ & $0.564$ & $0.619$ & $0.436$ \\
Attention MIL on frozen \textsc{DINOv2}~\citep{ilse2018attention} & $0.906$ & $0.574$ & $0.628$ & $0.332$ \\
\midrule
\methodA{} (simplex unmix), $30$ epochs   & $0.579 \pm 0.000$ & $0.660 \pm 0.001$ & $0.682 \pm 0.008$ & $-0.081 \pm 0.001$ \\
\methodB{} (proto match), closed-form     & $0.590 \pm 0.006$ & $\mathbf{0.683 \pm 0.016}$ & $\mathbf{0.722 \pm 0.011}$ & $\mathbf{-0.093 \pm 0.022}$ \\
\methodC{} (UFG channel-grouped), $30$ epochs & $\mathbf{0.689 \pm 0.008}$ & $0.635 \pm 0.043$ & $0.666 \pm 0.053$ & $+0.055 \pm 0.050$ \\
\midrule
\methodA{} (init only, $0$ epochs)        & $0.614$ & $0.657$ & $0.696$ & $-0.043$ \\
\methodB{} (init only $\equiv$ closed-form) & $0.599$ & $0.660$ & $0.708$ & $-0.062$ \\
\methodC{} (init only, $0$ epochs)        & $0.572$ & $0.654$ & $0.674$ & $-0.082$ \\
\bottomrule
\end{tabular}%
}
\end{table}

%% file: tables/tab_gigapath_lco.tex
\begin{table}[t]
\centering
\small
\caption{A larger encoder under the same LCO partition. Prov-GigaPath, the strongest encoder of the linear probe (Table~\ref{tab:encoder-probe}), replaces \textsc{DINOv2}-S/14 at seed $1337$, so its rows compare with the single-seed \textsc{DINOv2} rows here rather than with the three-seed means of Table~\ref{tab:bids-heldout}.}
\label{tab:gigapath-lco}
\begin{tabular}{llccc}
\toprule
\textbf{Encoder} & \textbf{Params} & \textbf{Decoder} & \textbf{In-dist val F1} $\uparrow$ & \textbf{Held-out F1} $\uparrow$ \\
\midrule
\multirow{3}{*}{\textsc{DINOv2} ViT-S/14} & \multirow{3}{*}{$22.1$ M} & \methodA{} & $0.5785$ & $0.6590$ \\
 & & \methodB{} & $0.5985$ & $0.6604$ \\
 & & \methodC{} & $0.6849$ & $0.6690$ \\
\midrule
\multirow{3}{*}{Prov-GigaPath ViT-G} & \multirow{3}{*}{$1135$ M} & \methodA{} & $0.5798$ & $0.6539$ \\
 & & \methodB{} & $0.6276$ & $0.6688$ \\
 & & \methodC{} & $0.7230$ & $0.6286$ \\
\midrule
\multirow{3}{*}{$\Delta$ ($51{\times}$ parameters)} & & \methodA{} & $+0.001$ & $-0.005$ \\
 & & \methodB{} & $+0.029$ & $+0.008$ \\
 & & \methodC{} & $+0.038$ & $-0.040$ \\
\bottomrule
\end{tabular}
\end{table}

%% file: tables/tab_headline.tex
\begin{table}[t]
\centering
\small
\caption{The decoders in distribution, on the random test split.}
\label{tab:headline}
\resizebox{\linewidth}{!}{%
\begin{tabular}{lccc}
\toprule
\textbf{Method} & \textbf{Per-sample F1} $\uparrow$ & \textbf{Macro F1} $\uparrow$ & \textbf{Exact match} $\uparrow$ \\
\midrule
\methodA{} (simplex unmix)              & $0.6086$ & $0.6259$ & $0.0374$ \\
\methodB{} (proto match)                & $0.6095$ & $0.6476$ & $0.0270$ \\
\methodC{} (UFG channel-grouped, $390$ params) & $\mathbf{0.6740}$ & $\mathbf{0.7011}$ & $\mathbf{0.0885}$ \\
\bottomrule
\end{tabular}%
}
\end{table}

%% file: tables/tab_illum_matched.tex
\begin{table}[t]
\centering
\small
\caption{Illumination correction in matched arms, where each decoder is trained \emph{and} evaluated under the same front-end, with or without the correction. Left: what the correction costs in distribution. Right: the largest per-sample F1 drop under each illumination corruption at its worst severity.}
\label{tab:illum-matched}
\resizebox{\linewidth}{!}{%
\begin{tabular}{lcc c cc cc cc}
\toprule
& \multicolumn{2}{c}{\textbf{Clean test F1}} & & \multicolumn{2}{c}{\textbf{Condenser gradient}} & \multicolumn{2}{c}{\textbf{Vignetting}} & \multicolumn{2}{c}{\textbf{Lamp intensity}} \\
\cmidrule{2-3} \cmidrule{5-6} \cmidrule{7-8} \cmidrule{9-10}
\textbf{Decoder} & with & without & & with & without & with & without & with & without \\
\midrule
\methodA{} & $0.6086$ & $0.6172$ & & $-0.0004$ & $-0.0061$ & $-0.0025$ & $-0.0065$ & $-0.0014$ & $-0.0020$ \\
\methodB{} & $0.6095$ & $0.6125$ & & $+0.0000$ & $-0.0742$ & $+0.0000$ & $\mathbf{-0.1356}$ & $-0.0090$ & $-0.0447$ \\
\methodC{} & $0.6740$ & $0.6826$ & & $-0.0005$ & $-0.0222$ & $+0.0000$ & $-0.0321$ & $-0.0224$ & $-0.0128$ \\
\bottomrule
\end{tabular}%
}
\end{table}

%% file: tables/tab_discovery.tex
\begin{table}[t]
\centering
\small
\caption{Discovery primitives under LOOCV, as mean $\pm$ std over the six folds. Cluster accuracy is recall times Hungarian-matched purity, and a single proposal at $K{=}1$ is pure by construction. $\Delta$F1 known is the change in known-class F1 after appending. Bold marks the best per column.}
\label{tab:discovery}
\resizebox{\linewidth}{!}{%
\begin{tabular}{l ccccc}
\toprule
\textbf{Discovery primitive} & $n_{\mathrm{proto}}$ & cluster acc $\uparrow$ & recall $\uparrow$ & purity $\uparrow$ & $\Delta$F1 known $\uparrow$ \\
\midrule
Greedy cosine ($\theta{=}0.7$, $m{\geq}50$) & $59.0$ & $0.443 \pm 0.12$ & $0.455 \pm 0.13$ & $0.977 \pm 0.03$ & $-0.377 \pm 0.09$ \\
\midrule
\textbf{Sinkhorn-Knopp ($K{=}1$)} & $\phantom{0}\mathbf{1.0}$ & $\mathbf{0.502 \pm 0.11}$ & $\mathbf{0.502 \pm 0.11}$ & $\mathbf{1.000 \pm 0.00}$ & $\mathbf{-0.031 \pm 0.01}$ \\
Sinkhorn-Knopp ($K{=}2$) & $\phantom{0}2.0$ & $0.397 \pm 0.12$ & $0.495 \pm 0.08$ & $0.789 \pm 0.12$ & $-0.083 \pm 0.02$ \\
Sinkhorn-Knopp ($K{=}4$) & $\phantom{0}4.0$ & $0.322 \pm 0.06$ & $0.567 \pm 0.14$ & $0.579 \pm 0.07$ & $-0.197 \pm 0.06$ \\
Sinkhorn-Knopp ($K{=}6$) & $\phantom{0}6.0$ & $0.338 \pm 0.08$ & $0.455 \pm 0.13$ & $0.757 \pm 0.10$ & $-0.245 \pm 0.08$ \\
\midrule
SK $K{=}1$ + Cr-KD distillation~\citep{gu2023class} & $\phantom{0}1.0$ & $0.110 \pm 0.19$ & $0.110 \pm 0.19$ & $1.000 \pm 0.00$ & $-0.030 \pm 0.01$ \\
\bottomrule
\end{tabular}%
}
\end{table}

%% file: sections/6_conclusion.tex
\section{Conclusion}
\label{sec:conclusion}

We introduced \phoebi{}, a wet-lab benchmark that asks two questions of every model. Does it recognise the species in a mixture it has never seen, and does it notice a species it has never met? The leave-combinations-out split tests only on mixtures absent from training.

Three findings stand out. First, classifiers that master familiar mixtures collapse on new ones whatever their backbone or pretraining, and the fault lies in how they pool evidence within an image, not in their features. Second, decoders anchored to each species do not collapse. They are stable straight from their closed-form prototypes, and the constraint that costs them accuracy in distribution is what protects them on new combinations. Third, the same frozen features serve the open world. A local nearest-neighbour score detects an unseen species where global scores cannot, and giving the new species its own prototype works best, leaving the known species almost untouched.

We expect the lesson to reach beyond microscopy. When a label is a set of parts that recombine, a model that reads each part against its own anchor holds steady on new combinations, and a model that learns the combinations does not. Neither question is solved yet, and \phoebi{} now measures progress on both questions with real bacterial cultures.

\textbf{Limitations.} \phoebi{} comes from one instrument, though the decoders resist illumination and detector shifts between microscopes. Its six rod-shaped species trade breadth for fine-grained difficulty, and a twelve-species, multi-instrument extension is planned. The open-world tasks hold out one species at a time, and each anchor needs a pure culture, which labs already keep.

\section{Broader Impact}
\label{sec:broader-impact}

\phoebi{} supports progress toward label-free phase-contrast microscopy as a screening instrument for mixed cultures, whose benefits would accrue most to under-resourced health centres, food and water safety pipelines and teaching laboratories where sequencing-based identification is unaffordable. The dataset contains no human subjects, personal data or clinical samples. \phoebi{} is a research benchmark and is not intended for clinical decision-making.

%% file: sections/x_supplementary.tex
\raggedbottom
\setcounter{topnumber}{3}
\setcounter{bottomnumber}{2}
\setcounter{totalnumber}{5}
\renewcommand{\topfraction}{0.92}
\renewcommand{\bottomfraction}{0.85}
\renewcommand{\textfraction}{0.06}
\renewcommand{\floatpagefraction}{0.72}
\setlength{\textfloatsep}{6pt plus 2pt minus 2pt}
\setlength{\floatsep}{5pt plus 2pt minus 2pt}
\setlength{\intextsep}{5pt plus 2pt minus 2pt}
\suppressfloats[t]

\begin{center}
\Large\textbf{PHOEBI: An Open-World Benchmark for
Multi-Label Bacterial Identification in Phase-Contrast
Microscopy}\\[0.5em]
\large (\textit{Supplementary Material})
\end{center}

\section{Dataset}
\label{app:dataset}

\subsection{Collection, Coverage and Labels}

\textbf{Dataset card.} Table~\ref{tab:dataset-card} summarises the dataset in one place, alongside the collection steps of Figure~\ref{fig:workflow} and the per-order and per-species statistics of Table~\ref{tab:dataset}.

\input{tables/tab_dataset_card}

\textbf{Combination coverage.} Tables~\ref{tab:coverage} and~\ref{tab:full-combos} give the coverage at each order and list the $40$ collected combinations. The gaps are deliberate and follow two design choices. First, multi-label targets saturate with order, since the all-present predictor already scores $0.80$ per-sample F1 at order $4$ and $0.91$ at order $5$. High-order combinations therefore separate models poorly, and that culture budget went to orders $2$ and $3$ instead. Second, we avoided mixing \texttt{bs} and \texttt{bt}, the only same-genus pair and the hardest to tell apart by morphology. Only $2$ of the $11$ possible combinations containing both were collected, which accounts for $9$ of the $17$ gaps at orders $2$ to $4$. Each species alone remains well covered (\texttt{bs} in $17$ combinations, \texttt{bt} in $13$ and the other four in $18$ or $19$), so the constraint limits co-occurrence rather than coverage. The three missing pairs are \texttt{bt\_fj}, \texttt{ka\_pf} and \texttt{bs\_bt}.

\input{tables/tab_coverage}

\begin{table}[ht]
\centering
\small
\caption{All $40$ species combinations in the \phoebi{} dataset.}
\label{tab:full-combos}
\begin{tabular}{rlr|rlr}
\toprule
\textbf{\#} & \textbf{Tokens} & \textbf{Order} & \textbf{\#} & \textbf{Tokens} & \textbf{Order} \\
\midrule
 1 & \texttt{bs}               & 1 & 21 & \texttt{bs\_mx\_pf}       & 3 \\
 2 & \texttt{bt}               & 1 & 22 & \texttt{bs\_ka\_fj}       & 3 \\
 3 & \texttt{mx}               & 1 & 23 & \texttt{bs\_ka\_pf}       & 3 \\
 4 & \texttt{ka}               & 1 & 24 & \texttt{bs\_fj\_pf}       & 3 \\
 5 & \texttt{fj}               & 1 & 25 & \texttt{bt\_mx\_ka}       & 3 \\
 6 & \texttt{pf}               & 1 & 26 & \texttt{bt\_mx\_pf}       & 3 \\
 7 & \texttt{bs\_mx}           & 2 & 27 & \texttt{bt\_ka\_fj}       & 3 \\
 8 & \texttt{bs\_ka}           & 2 & 28 & \texttt{bt\_fj\_pf}       & 3 \\
 9 & \texttt{bs\_fj}           & 2 & 29 & \texttt{mx\_ka\_fj}       & 3 \\
10 & \texttt{bs\_pf}           & 2 & 30 & \texttt{mx\_ka\_pf}       & 3 \\
11 & \texttt{bt\_mx}           & 2 & 31 & \texttt{mx\_fj\_pf}       & 3 \\
12 & \texttt{bt\_ka}           & 2 & 32 & \texttt{ka\_fj\_pf}       & 3 \\
13 & \texttt{bt\_pf}           & 2 & 33 & \texttt{bs\_bt\_mx}       & 3 \\
14 & \texttt{mx\_ka}           & 2 & 34 & \texttt{bs\_mx\_ka\_pf}   & 4 \\
15 & \texttt{mx\_fj}           & 2 & 35 & \texttt{bt\_mx\_ka\_fj}   & 4 \\
16 & \texttt{mx\_pf}           & 2 & 36 & \texttt{bs\_bt\_ka\_fj}   & 4 \\
17 & \texttt{ka\_fj}           & 2 & 37 & \texttt{bs\_mx\_fj\_pf}   & 4 \\
18 & \texttt{fj\_pf}           & 2 & 38 & \texttt{bt\_ka\_fj\_pf}   & 4 \\
19 & \texttt{bs\_mx\_ka}       & 3 & 39 & \texttt{bs\_ka\_fj\_pf}   & 4 \\
20 & \texttt{bs\_mx\_fj}       & 3 & 40 & \texttt{bs\_bt\_mx\_ka\_fj\_pf} & 6 \\
\bottomrule
\end{tabular}
\end{table}

\textbf{Culture protocol and label verification.}\phantomsection\label{app:label-noise} The six species are never co-cultured. A trained microbiologist grows each one separately to its characteristic stage and checks under the microscope that it is present and morphologically correct. A combination is made only at acquisition time, by mixing the verified pure suspensions at a controlled volume ratio, and the mixture goes straight onto a plain glass slide as a wet mount with no staining, fixation, embedding or mountant. All $40$ combinations share this single mounting step. Three consequences follow. No species can be lost to competition during growth, because there is no shared growth phase. Every species in a combination was confirmed in the material that went onto the slide, so labels are verified at culture level. And slide preparation leaves little to vary between combinations, so it cannot serve as a shortcut cue. What a culture-level label cannot promise is presence in every field, since a species may miss a given $1024 \times 1024$ field by sampling. That is a property of any finite field of view rather than of the annotation.

Table~\ref{tab:label-noise} bounds this rate without manual annotation. In a pure culture the labelled species is in every field by construction, so a false negative there is model error. In a mixture a species may truly be absent from a field. The gap between the two false-negative rates under \methodB{} therefore bounds how often a culture-level label overstates presence in a field, at $4.9\%$ on average. The bound is conservative, because the extra difficulty of reading mixtures also sits inside the gap.

\input{tables/tab_label_noise}

\subsection{Colour}
\label{app:rgb-cast}

\begin{figure}[ht]
    \centering
    \includegraphics[width=\linewidth]{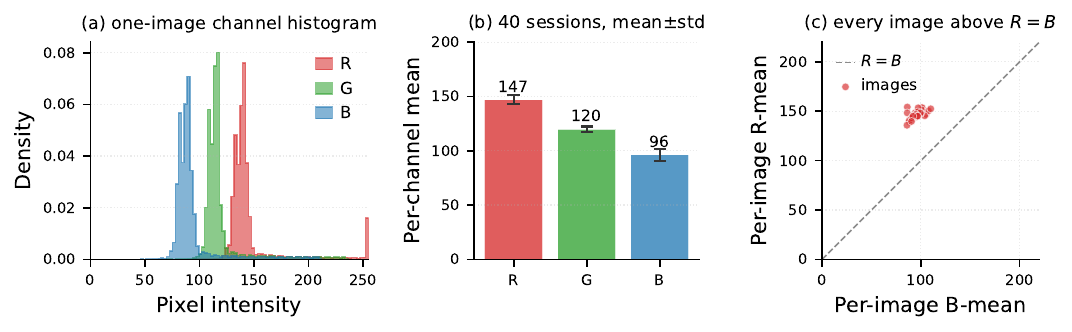}
    \caption{The colour camera gives true RGB images with the same warm cast in every session. The dashed line in (c) marks where the red and blue means are equal.}
    \label{fig:channel-cast}
\end{figure}

The microscope's Bayer colour camera records true three-channel images with a strong and reproducible warm cast (Figure~\ref{fig:channel-cast}). Across one image from each of the $40$ culture sessions the channel means are $(R, G, B) = (147,\, 120,\, 96)$, and every image lies above the $R = B$ line. The cells are unstained, so the colour comes entirely from the lamp's colour temperature and the camera's spectral response. All three channels feed \textsc{DINOv2} unchanged, and the illumination correction runs per channel, which removes the spatial hotspot and keeps the relative cast.

The cast is a uniform property of the acquisition rather than a species cue. We converted $40$ pure-culture test images per species to luma, replicated it to three channels and re-scored the images with the frozen RGB-trained \methodB{}, on one feature path so that every comparison is paired (Table~\ref{tab:grayscale}). Similarity to each species' own prototype falls by a near-constant amount, while the margin over the best competing prototype does not fall for any species. Colour therefore carries no species signal in this data. It carries a consistent offset that the released model was calibrated on, which is why grayscale inputs fall below the thresholds for four of the six species. The pipeline keeps all three channels, which the instrument produces with the same offset in every session.

\input{tables/tab_grayscale}

\subsection{Datasheet}
\label{app:datasheet}

\phoebi{} follows the structured form of~\citet{gebru2021datasheets}.

\textbf{Motivation, composition, collection.} \phoebi{} supports open-world multi-label bacterial recognition under conditions close to clinical and environmental practice, where mixed cultures are the norm, new species are routine and morphology is fine-grained. Existing benchmarks are mostly single-label and closed-set, so \phoebi{} trades species count for combinatorial coverage and pairs the data with a leave-combinations-out protocol. The authors created it at the University of Central Florida. Each instance is a $1024 \times 1024 \times 3$ RGB JPEG phase-contrast image with a six-dimensional binary label vector taken from its folder name. The release holds about $120{,}000$ images of the $40$ combinations, each imaged across several acquisition sessions with several crops per session. Crops from one session are correlated, which is why the leave-combinations-out split holds out whole combinations and the in-distribution split follows acquisition order rather than a shuffle. Coverage, its design constraints and the culture protocol are in Appendix~\ref{app:dataset}.

\textbf{Splits, processing, scope.} Two deterministic splits are released. The in-distribution $80/10/10$ split is assigned in acquisition order within each combination by \texttt{tools/\allowbreak build\_splits.py}, and the leave-nine-combinations-out split is drawn at seed $1337$ by \texttt{baselines/\allowbreak supervised\_\allowbreak multilabel\_\allowbreak heldout.\allowbreak select\_\allowbreak heldout()}. The development folds of Appendix~\ref{app:devprotocol} are derived from the latter by \texttt{tools/\allowbreak build\_lco\_\allowbreak dev\_\allowbreak folds.py}. Labels give the composition set by the experiment and encoded in each folder name. Every pure culture was checked before mixing, and presence in individual fields is not annotated, though Appendix~\ref{app:label-noise} bounds how often it departs from the culture-level label. \phoebi{} supports presence detection, compositional generalization, open-set rejection and novel-class discovery. It is not intended for clinical decision-making and does not support proportion estimation. A twelve-species, multi-instrument extension is planned as version 2.0.

\textbf{Access and reproducibility.} The dataset, with all $120{,}000$ images, the random and leave-combinations-out split manifests, a Croissant metadata file and a CC-BY-4.0 licence, is at \href{https://huggingface.co/datasets/sochastic/PHOEBI}{\nolinkurl{huggingface.co/datasets/sochastic/PHOEBI}}. The complete pipeline, from the tiling and illumination front-end through the three decoders, every baseline and the LCO and LOOCV drivers to the batch scripts behind each table, is at \href{https://github.com/eternal-f1ame/phoebi}{\nolinkurl{github.com/eternal-f1ame/phoebi}}, and the project page at \href{https://phoebi-benchmark.vercel.app}{\nolinkurl{phoebi-benchmark.vercel.app}} links both. The released selection procedure regenerates the nine held-out combinations of Section~\ref{subsec:protocol} at seed $1337$, and seeds $1338$ and $1339$ give the other two partitions behind every mean $\pm$ std. The development folds are listed in Appendix~\ref{app:devprotocol} and rebuilt from the partition at the same seed. Decoder equations, training and calibration are in Appendix~\ref{app:decoder-details}, and the open-set and discovery harness is Algorithm~\ref{alg:loocv}. The corruption battery and the matched illumination arms are specified in Appendix~\ref{app:acquisition}, and the encoder-capacity control in Appendix~\ref{app:gigapath}.

\section{Pipeline}
\label{app:pipeline}

\subsection{Front-End and Decoders}
\label{app:decoder-details}

\textbf{Tile pipeline.}\phantomsection\label{app:tile-pipeline} For an image $I: \Omega \to \R^3$ we estimate a per-channel background $B_c = G_\sigma \ast I_c$ via a large-$\sigma$ Gaussian and form $\tilde{I}_c = I_c / (B_c / \bar{B})$, picking $\sigma = 64$ px so cellular structure ($5$ to $20$ px) is preserved while the lamp gradient (hundreds of px) is captured. Positional embeddings are interpolated to a $16{\times}16$ patch grid for the $s = 224$ tiles of Section~\ref{subsec:approach}.

\textbf{\methodA{}: equations and training.}\phantomsection\label{app:simplex-details} Let $\proto \in \R^{K \times D}$ be a matrix of L$_2$-normalized class prototypes. For a tile $z_t$ we compute scaled cosine logits, project them onto the probability simplex via sparsemax~\citep{martins2016sparsemax}, reconstruct the tile, and record the per-tile residual:
\begin{align}
\ell_t = \tau \proto z_t \in \R^K, \quad
w_t = \sparsemax(\ell_t) \in \simplex, \quad
\hat{z}_t = \proto^\top w_t, \quad
r_t = z_t - \hat{z}_t.
\end{align}
The image-level read-outs are $w(x) = \tfrac{1}{T}\sum_t w_t$ for presence, predicting $\hat{y}_k = \ind[w_k(x) > \theta_k^A]$ with $\theta_k^A$ the $5$th percentile of $w_k$ over positive validation images, and $r(x) = \tfrac{1}{T}\sum_t \lVert r_t\rVert$ for the residual norm. Training minimizes the reconstruction error $\E_x \tfrac{1}{T}\sum_t \lVert z_t - \hat{z}_t\rVert_2^2$ over the prototype matrix for thirty epochs. We omit entropy regularization, because it pushes $w_t$ toward uniform and cancels the structural sparsity. Both decoders start from the pure-culture means $\proto_k = \tfrac{1}{N_k}\sum_{i: y_k=1} z_i / \lVert\cdot\rVert_2$, which \methodB{} keeps as they are and \methodA{} refines.

\textbf{Mathematical foundations.}\phantomsection\label{app:math} Two properties of the front-end carry design weight. Sparsemax~\citep{martins2016sparsemax} is the Euclidean projection onto $\simplex$, so absent species receive exact zeros rather than small positive mass. Its Jacobian $\mathrm{diag}(s) - s s^\top / \lVert s \rVert_1$, with $s$ the support indicator, passes gradient only through active components, so an absent species contributes nothing to the prototype update. And under Assumption~H the tiles of an image are i.i.d.\ draws from $\Prob(\cdot \mid y(x))$, so every per-tile statistic we aggregate, whether sparsemax weight, cosine similarity, residual norm or maximum similarity, has mean-aggregation variance $\mathcal{O}(1/T)$. Figure~\ref{fig:tile-curve} checks that rate empirically, and Figure~\ref{fig:spatial-homogeneity} supports the i.i.d.\ premise itself at both the pixel and the embedding level.

\subsection{Ablations}
\label{app:ablations}

\textbf{Encoder probe across pretraining objectives.}\phantomsection\label{app:encoder-probe} All thirteen encoders in Table~\ref{tab:encoder-probe} share the same tile and illumination pipeline, a single $\texttt{nn.Linear}(D, K)$ BCE head and per-class thresholds calibrated on the validation split. Two of the four biomedical models rise above the band of the nine general-purpose backbones, namely Prov-GigaPath~\citep{xu2024gigapath}, pretrained on $1.3$B histopathology tiles, and Phikon~\citep{filiot2024phikon}, while UNI~\citep{chen2024uni} and BiomedCLIP~\citep{zhang2023biomedclip} fall inside it. The ordering fits modality transfer, since biomedical pretraining rewards sensitivity to texture, sub-cellular structure and aspect ratio that natural-image pretraining smooths over. It is also immaterial to the compositional result, because the six points between the best and worst encoder are an order of magnitude smaller than the $0.39$ to $0.57$ collapse.

\textbf{Encoder capacity.}\phantomsection\label{app:gigapath} A separate question is whether a stronger encoder closes the compositional gap. We re-ran the full LCO protocol on the seed-$1337$ partition with Prov-GigaPath (ViT-G, $1135$M parameters, the probe's best encoder) in place of \textsc{DINOv2}-S/14 ($22.1$M) under all three decoders (Table~\ref{tab:gigapath-lco}). The larger encoder raises validation F1 for every decoder and keeps held-out F1 within a point for \methodA{} and \methodB{}, while \methodC{} loses $4.0$ points held-out and its $\Delta$F1 widens from $+0.016$ to $+0.094$. Encoder capacity buys in-distribution accuracy and not compositional generalisation, the conclusion the probe reaches from the other direction.

\textbf{Tile size and illumination.} \methodB{} prefers smaller tiles, and both decoders degrade above $s = 224$ (Table~\ref{tab:ablations}). We keep $s = 224$ as the default because it matches \textsc{DINOv2}'s training resolution and avoids the memory overhead of non-divisible tiles at $s = 168$. The small in-distribution cost of illumination correction buys the optical invariance measured in Appendix~\ref{app:acquisition}. \methodB{} is the decoder that depends on it, as its construction predicts, since pure cosine similarity to fixed prototypes is the read-out most exposed to a global intensity shift.

\input{tables/tab_ablations}

\textbf{Sparsemax against softmax in \methodA{}.} Replacing sparsemax with softmax removes the structural zeros. On the random test split softmax gains a point of per-sample F1 (Table~\ref{tab:ablations}) and $2.1$ points of macro F1 ($0.650$ against $0.630$) but loses $1.2$ points of exact match ($0.024$ against $0.036$). Softmax spreads small positive weights over absent classes, which the per-class thresholds absorb, whereas the exact zeros of sparsemax are what exact match rewards and what the residual is defined against. Sparsemax stays for that structural role rather than for in-distribution F1.

\textbf{Boundary tiles and Assumption H.} To test whether H breaks at the edge of the field, we re-ran \methodB{} on the LCO protocol with only the central $2{\times}2$ tiles of the $4{\times}4$ grid, which excludes every tile that touches the image edge, and recalibrated its thresholds on validation for that tile set. Held-out F1 changes by $-0.012$ ($0.660 \to 0.648$), validation F1 by $-0.003$ and held-out macro F1 by $-0.017$. The tile-count sweep predicts a drop of $0.018$ for \methodB{} from $T = 16$ to $T = 4$ (Figure~\ref{fig:tile-curve}), so the twelve edge tiles are worth as much as interior ones. H holds at the edge of the field, and the pipeline keeps the full grid for the variance reduction that sixteen tiles bring.

\textbf{Isotonic recalibration for \methodB{}.} We compared three threshold protocols on the test split. The $5$th-percentile validation thresholds on raw similarities give an F1 of $0.6095$, argmax-F1 thresholds on raw similarities give $0.6174$, and argmax-F1 thresholds on isotonic-calibrated scores also give $0.6174$. Isotonic calibration therefore changes nothing once the threshold rule is matched, and its $0.79$-point gain over the deployed rule comes from the rule alone. The pipeline does not deploy it and recommends it only when calibrated probabilities are needed.

\textbf{Asymmetric loss and tile-level Mixup for \methodC{}.} Asymmetric loss~\citep{ridnik2021asl} ($\gamma_- = 4$, margin shift $0.05$) and tile-level Mixup~\citep{zhang2018mixup} ($\alpha = 0.4$, with the multi-label union taken as an element-wise max, which H justifies) replace BCE as drop-in changes. Test F1 is $0.674$ for BCE, $0.671$ for asymmetric loss, $0.673$ for BCE with Mixup and $0.671$ for asymmetric loss with Mixup. All four sit within $0.003$ of one another, so \methodC{} at $K = 6$ on frozen \textsc{DINOv2} features is not limited by its training loss.

\subsection{Acquisition Robustness}
\label{app:acquisition}

\textbf{Corruption battery.} Every image in \phoebi{} comes from one instrument. To probe the acquisition variables that differ most between microscopes without a second instrument, we inject each corruption into the raw $8$-bit frame before the illumination correction, where a real acquisition artefact would arise, and pass the result through the same feature path behind every published number, with the frozen decoders and thresholds untouched. Seven corruptions at three severities each cover the illumination path and the non-optical failures. A multiplicative ramp across the field stands in for condenser misalignment, a radial falloff for vignetting, a global intensity scale for lamp voltage or exposure and a gamma curve for the detector response. Gaussian blur stands in for focus drift, additive Gaussian noise for a different camera or gain and JPEG re-encoding for capture software. An eighth condition resizes the whole $1024$ px field to a single $224$ px tile in place of the $4{\times}4$ grid. The caption of Table~\ref{tab:corruptions} lists the parameters. The sweep runs on a $4{,}000$-image stratified subsample of the test split whose clean F1 matches the full split to within $0.001$, and the clean arm on the full split reproduces Table~\ref{tab:headline} exactly, which ties the battery to the published numbers.

\input{tables/tab_corruptions}

Table~\ref{tab:corruptions} reports the worst case over the three severities. The three illumination corruptions and the gamma curve cost any decoder at most $2.2$ points, and condenser misalignment and vignetting at most $0.25$. Focus drift, sensor noise and JPEG re-encoding are the corruptions that matter, and \methodA{} is the most robust decoder under each, because the simplex projection discards small perturbations of the tile embedding that a raw cosine read-out passes straight through. Tiling is essential, since a single resized tile costs $21$ to $30$ points of per-sample F1.

\textbf{Matched illumination arms.} The battery keeps the correction in place. To measure what the correction costs and what it buys, we retrained and recalibrated all three decoders without it and evaluated them under that same front-end (Table~\ref{tab:illum-matched}). In distribution the correction costs $0.30$ to $0.86$ points. Under the illumination corruptions \methodA{} and \methodC{} stay fairly robust either way, while \methodB{} without the correction loses $7.4$ points under a condenser gradient and $13.6$ under vignetting, against a worst case of $0.9$ with it. For \methodB{} the correction is a $0.30$-point price in distribution for $13.6$ points of robustness.

\section{Compositional Generalization}
\label{app:experiments}

\subsection{Uncertainty and Replication}
\label{app:devprotocol}

\textbf{Combination-level uncertainty.} Every image of a combination carries the same label vector, so for any species the held-out positives and negatives are unions of whole combinations, and every held-out per-species metric compares combinations. On the seed-$1337$ partition, $63$ to $90\%$ of a decoder's per-species score variance lies between combinations, and the AUROC computed on the nine combination means ($0.595$) matches the image-level value ($0.606$). The nine held-out combinations are therefore reported once, with intervals, and model development uses held-out combinations of its own. The released folds split the $26$ trained-on combinations of orders $2$ to $4$ into five folds (Table~\ref{tab:devfolds}). Each fold is held out once and the singletons stay in training. Because the partition already holds out the singleton \texttt{bt}, one species lacks a pure culture in every fold, just as at test time. In each fold the decoders train on the training images of the remaining combinations, calibrate on their validation images and are evaluated on all images of the held-out development combinations. Predictions are pooled over folds, and every metric carries a $95\%$ bootstrap interval over combinations. Selection ranks configurations by pooled mean per-species AUROC first and by macro-F1 margin over the all-present predictor second. Per-sample F1 alone is not a criterion, because the all-present predictor already reaches $0.80$ on every order-$4$ combination.

\begin{table}[ht]
\centering
\small
\caption{Development folds for the seed-$1337$ partition (\texttt{lco\_dev\_folds.json}).}
\label{tab:devfolds}
\begin{tabular}{cl}
\toprule
\textbf{Fold} & \textbf{Held-out development combinations} \\
\midrule
$0$ & \texttt{bs\_fj}, \texttt{fj\_pf}, \texttt{bt\_ka\_fj}, \texttt{bt\_mx\_pf}, \texttt{mx\_ka\_pf}, \texttt{bt\_mx\_ka\_fj} \\
$1$ & \texttt{bt\_ka}, \texttt{bt\_mx}, \texttt{bs\_mx\_pf}, \texttt{bt\_mx\_ka}, \texttt{ka\_fj\_pf} \\
$2$ & \texttt{bt\_pf}, \texttt{mx\_fj}, \texttt{bs\_fj\_pf}, \texttt{bs\_ka\_fj}, \texttt{bs\_ka\_fj\_pf} \\
$3$ & \texttt{bs\_mx}, \texttt{mx\_ka}, \texttt{bs\_bt\_mx}, \texttt{bt\_fj\_pf}, \texttt{bt\_ka\_fj\_pf} \\
$4$ & \texttt{bs\_ka}, \texttt{mx\_pf}, \texttt{bs\_mx\_ka}, \texttt{mx\_fj\_pf}, \texttt{bs\_mx\_ka\_pf} \\
\bottomrule
\end{tabular}
\end{table}

\input{tables/tab_intervals}

Table~\ref{tab:intervals} applies these intervals to the three decoders, pooling the $27$ held-out combinations of the three partitions (resampling partition and combination pairs, $1{,}000$ resamples) and, separately, under the development protocol with one training per fold. Under the protocol \methodB{} and \methodA{} sit at the all-present predictor and separate only \texttt{ka}. \methodC{} falls measurably \emph{below} it, with a per-sample F1 margin of $-0.077$ $[-0.132, -0.023]$, and inverts the ranking of \texttt{fj}, which is the decoder-level form of the collapse with an interval around it. The development folds hold out orders $2$ to $4$ only, whereas every test partition also holds out a singleton and the six-species mixture, whose images sit at the extremes of every species' score distribution. Absolute development and test numbers are therefore not comparable, and only the ordering of decoders carries over. Across $36$ closed-form configurations (six unit granularities, two read-outs and three aggregations) the development AUROC lies between $0.51$ and $0.60$ with overlapping intervals, so no unit differs detectably from the paper's choice.

\textbf{Replication on the four-class subset.}\phantomsection\label{app:4class} Table~\ref{tab:supervised-baselines-4class} repeats the nine-backbone end-to-end fine-tuning sweep on the four-class subset, whose $14$ combinations use folder codes \texttt{b}, \texttt{f}, \texttt{k} and \texttt{p} for \texttt{bs}, \texttt{fj}, \texttt{ka} and \texttt{pf}. It is a separate culture batch imaged in a separate session on the same instrument, so the replication spans sessions but not instruments. The training recipe is identical, and the held-out split removes one single species, two pairs, one triple and the only quadruple while keeping every species in training through another combination. Rankings of individual backbones change with the held-out set, while the collapse itself is uniform across all nine.

\subsection{Decoder Behaviour}

\textbf{Held-out F1 by combination order.}\phantomsection\label{app:per-order} Held-out F1 rises steadily with combination order (Figure~\ref{fig:per-order-f1}), and so does the all-present predictor, whose per-sample F1 at order $k$ is $2k/(k+6)$. From order $2$ upward the decoders stay within $0.12$ of that curve, which is why the order composition of the held-out set sets the sign of $\Delta$F1 in Table~\ref{tab:bids-heldout}. Singletons are the hardest case. With no pure culture of \texttt{bt} available at calibration, \methodA{} and \methodB{} land just above the all-present predictor's $0.29$ ($0.296 \pm 0.025$ and $0.322 \pm 0.167$ over the three seeds), and \methodC{} drops to near zero ($0.019 \pm 0.025$), because a head initialised from mixed cultures starts in the wrong region. The two geometric anchors differ in stability rather than level. \methodA{}'s simplex projection denoises the mixed-culture initialisation to the same answer on every seed, while \methodB{}'s raw cosine read-out inherits its direction, and its seed-to-seed spread is nearly seven times larger.

\textbf{Init-only against trained, all three decoders.} Setting the training epochs of \methodA{} and \methodC{} to $0$ and re-running the LCO protocol separates what each decoder gains in distribution from what it gains on new combinations (bottom block of Table~\ref{tab:bids-heldout}). \methodC{}'s init-only row coincides with the all-present predictor, since an untrained head under argmax-F1 thresholds predicts every species. Its validation gain from training is therefore entirely learned, and its held-out score starts from the constant reference. Training the simplex-constrained \methodA{} for $30$ epochs gives up $3.5$ points of validation F1 and gains $0.3$ points held-out, within seed noise. The gradient refines prototypes initialised from mixed cultures, and the structural sparsity of the simplex projection prevents per-combination overfitting. \methodB{} has nothing to train, so its init-only row is its seed-$1337$ result. Training the unanchored \methodC{} for $30$ epochs gives the opposite trade, gaining $11.7$ points of validation F1 and losing $1.9$ held-out. The channel-grouped head is structurally a per-image classifier on $D/K = 64$-dimensional subspaces, so without an anchor the gradient builds per-combination decision surfaces, the mechanism behind the collapse in Table~\ref{tab:supervised-baselines}.

\textbf{Threshold calibration and per-species error rates.}\phantomsection\label{app:calibration} Presence thresholds for \methodA{} and \methodB{} are set at the $5$th percentile of each class's score over positive validation images, which deliberately favours recall, since in open-world identification a missed species costs more than an extra one. The asymmetry shows clearly on the held-out set. For \methodA{}, five of the six thresholds fall to $\theta_k^A \le 0.05$, because the simplex weight of each species' lookalike is non-zero on virtually every image. This gives a false-negative rate (FNR) of at most $0.08$ and an FPR between $0.95$ and $0.97$ for \texttt{bs}, \texttt{fj}, \texttt{mx} and \texttt{pf}. Only \texttt{ka} is calibrated tightly ($\theta_k = 0.10$, FNR $0.02$, FPR $0.30$), since its encapsulated short cells give a well-separated score distribution. \methodB{} repeats the pattern at higher absolute thresholds on raw cosine similarities. \methodC{} is the most discriminative. Its BCE-trained logits give a lower FPR on \texttt{ka} ($0.05$) and \texttt{bs} ($0.58$), at the cost of a markedly higher FNR on \texttt{bt} ($0.67$) and on \texttt{fj} ($0.38$).

Under \methodA{} the mean score of \texttt{bt}-positive images even falls \emph{below} that of \texttt{bt}-negative images ($-0.02$). This follows from the geometry behind the per-class F1 of Figure~\ref{fig:per-class-f1}. The \texttt{bt} prototype absorbs almost no sparsemax mass beyond what the \texttt{bs} and \texttt{mx} prototypes already explain, so the routing gives \texttt{bt} near-zero weight whether or not it is present, and the mean over tiles is dominated by the other species of the mixture. The loose threshold ($\theta = 0.05$ rather than the ROC-optimal value) still recovers a per-sample F1 of $0.692$ on the three held-out combinations that contain \texttt{bt}, and tightening it trades recall for precision. On the seed-$1337$ partition \methodB{} shows the same asymmetry. It rejects \texttt{ka} reliably (FPR $0.04$) while the other five species carry FPR $0.89$ to $0.99$, so on $56\%$ of held-out images its prediction coincides with the all-present predictor. \methodC{}, calibrated by argmax-F1, predicts $3.9$ species per image with FPR between $0.01$ and $0.50$ on four species, at a recall of only $0.69$ on \texttt{bs} and $0.37$ on \texttt{bt}. Pooled over the three partitions ($27$ held-out combinations), the per-species AUROC intervals exclude chance only for \texttt{ka} with every decoder, for \texttt{fj} with \methodB{} ($0.66$ $[0.51, 0.81]$) and for \texttt{mx} with \methodC{} ($0.73$ $[0.54, 0.89]$). \methodA{} ranks \texttt{mx}-containing combinations below \texttt{mx}-free ones ($0.32$ $[0.19, 0.47]$), and \texttt{bs}, \texttt{bt} and \texttt{pf} stay unresolved for every decoder.

\begin{figure}[ht]
    \centering
    \includegraphics[width=\linewidth]{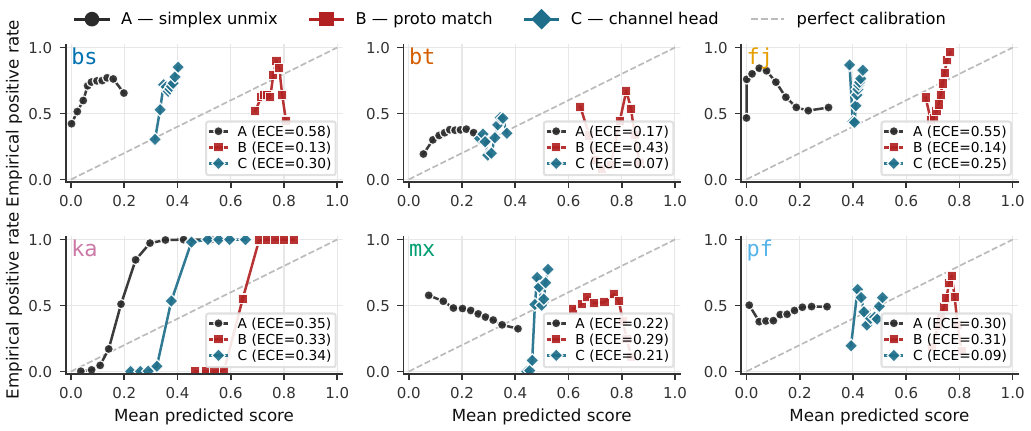}
    \caption{Per-species reliability diagrams on the LCO held-out set, with the expected calibration error (ECE) of each decoder in the legends. Points below the diagonal over-predict presence.}
    \label{fig:reliability}
\end{figure}

\textbf{Reliability.} None of the three decoders is well calibrated under LCO shift, but each fails differently (Figure~\ref{fig:reliability}). \methodA{}'s simplex weights over-predict the easy classes, \methodB{}'s raw cosine similarities give a near-flat curve because the scores are compressed, and \methodC{}'s BCE logits are the best calibrated for most species but inherit the \texttt{bt} failure. The decision rule ranks species rather than estimating posteriors, so a user who needs calibrated probabilities should recalibrate each class isotonically on a held-out validation slice.

\begin{figure}[ht]
    \centering
    \includegraphics[width=\linewidth]{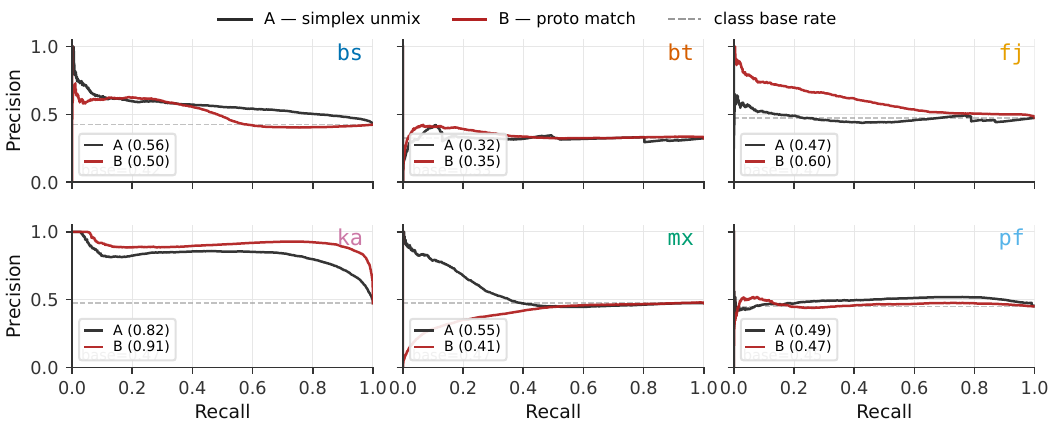}
    \caption{Per-class precision-recall curves on the random test split for \methodA{} and \methodB{}, with average precision in each legend and the class base rate dashed. \methodC{} is omitted because its per-tile scores were not kept.}
    \label{fig:pr-curves}
\end{figure}

\textbf{Per-class precision and recall.} The precision-recall curves rank the classes as the per-class F1 of Figure~\ref{fig:per-class-f1} does (Figure~\ref{fig:pr-curves}). \texttt{ka} is the only class that either decoder holds above $0.8$ precision over a non-trivial range of recall, and the other five stay close to their base rates.

\textbf{Failure modes.} The same difficulty ordering, $\texttt{bt} \prec \texttt{bs} \approx \texttt{fj} \approx \texttt{mx} \approx \texttt{pf} \prec \texttt{ka}$, appears in every evaluation. \texttt{bt} cells are thin rods close in shape to \texttt{bs} and \texttt{mx}, so a \texttt{bt} tile embeds within the span of the \texttt{bs} and \texttt{mx} prototypes, while the large, encapsulated short rods of \texttt{ka} stand apart and stay easy throughout. This geometry is why the residual fails when \texttt{bt} is held out. A held-out \texttt{bt} tile is reconstructed with a small residual by a sparse mix of \texttt{bs} and \texttt{mx}, just like the known \texttt{bs} and \texttt{mx} tiles, and the residual AUROC on that fold drops below chance ($0.390$). The $k$-NN score separates them, because the cosine distance to the local neighbourhood still tells held-out \texttt{bt} from known \texttt{bs} and \texttt{mx}. Its gain of $0.298$ AUROC on the \texttt{bt} fold, to $0.688$, exceeds its mean lift of $0.257$ across folds. The narrow spread of the encoder probe (Table~\ref{tab:encoder-probe}) makes the ordering a property of the species' geometry under phase contrast rather than of \textsc{DINOv2}. Exact match falls below $0.10$ on quadruples and above for every decoder, an artefact of independent per-class thresholds, since at a per-class F1 of $0.80$ six-class exact match is bounded by $0.80^6 \approx 0.26$. It therefore serves only as a side metric.

\subsection{Decoder Variants}

\textbf{Ensembling \methodA{} and \methodB{}.} Averaging the normalised scores of the two decoders, which share one prototype representation, does not improve on \methodA{} alone. The best ensemble, a soft OR at threshold $0.40$, reaches a per-sample F1 of $0.614$ against \methodA{}'s $0.660$. The two decoders are positively correlated, both scoring high on \texttt{ka} and low on \texttt{bt}, so averaging them adds no independent signal.

\textbf{Per-class learnable temperature.}\phantomsection\label{app:method-a-ablations} Two further ablations of \methodA{} target the confusion among \texttt{bt}, \texttt{bs} and \texttt{mx} seen in Figure~\ref{fig:per-class-f1}, on the LCO protocol over seeds $1337$ to $1339$. The default \methodA{} scales prototype-feature cosine similarities into sparsemax logits with one fixed $\tau = 10$. Replacing it with six learnable log-temperatures, one per prototype and trained with the prototypes through the reconstruction loss, lifts validation F1 from $0.578 \pm 0.001$ to $0.593 \pm 0.001$ and held-out F1 from $0.659 \pm 0.000$ to $0.679 \pm 0.002$, and leaves $\Delta$F1 almost unchanged ($-0.086$ against $-0.081$). The anchor is set by the prototype directions rather than by the routing scale, so a per-class $\tau$ refines the logits without changing where the simplex weights fall on new combinations.

\textbf{Hyperspherical prototype repulsion.} Adding $\lambda \cdot \mathrm{mean}_{i \neq j}\, \exp(P_i \cdot P_j / \tau_r)$ to the reconstruction loss ($\tau_r = 0.1$, $\lambda \in \{0, 0.01, 0.1, 1.0\}$) pushes the prototypes apart on the unit sphere without changing inference, with the aim of moving the \texttt{bt} prototype out of the span of \texttt{bs} and \texttt{mx}. The repulsion works geometrically, since the mean off-diagonal prototype similarity falls from $0.652$ at $\lambda = 0$ to $-0.194$ at $\lambda = 1.0$. Held-out F1 falls with it, from $0.659$ to $0.490$, $0.395$ and $0.356$ as $\lambda$ grows. The \texttt{bt} prototype truly belongs \emph{between} those of \texttt{bs} and \texttt{mx}, because \texttt{bt} cells are morphologically intermediate, so a penalty on proximity drives it off the data manifold. \texttt{bt} is hard because of its morphology rather than because the prototypes sit too close, and the deployed \methodA{} handles it through the denoising of the simplex projection rather than through prototype geometry.

\textbf{Finer units, learned subspaces and combination coverage.} A threshold-free held-out AUROC near $0.5$ means the pooled CLS embedding of a $224$ px tile is not additive over the species it contains. We therefore tested whether changing the unit that the simplex unmixes restores discrimination, with the backbone frozen throughout. Closed-form prototypes at six granularities (CLS, patch-mean and per-patch units at $224$ px, CLS and patch-mean at $112$ px, and CLS at $56$ px) with mean, $90$th-percentile and max aggregation span $0.59$ to $0.63$ held-out AUROC against the paper's $0.61$, and smaller tiles are no better than $224$ px. Training the prototypes on $4.7$M patch tokens with the reconstruction objective plateaus within two epochs. Adding a penalty on the sparsemax mass given to absent species, which Assumption H licenses, reaches $0.68$ at $\lambda = 10$ by moving discrimination between confusable species (\texttt{mx} up, \texttt{fj} down) while doubling the reconstruction loss. At $\lambda = 1$ its thresholded numbers ($0.688$ and $0.721$) match the \methodB{} row within seed noise. An orthonormal projection learned under the simplex constraint ($d \in \{32, 64, 128\}$) fits the training distribution, with development AUROC up to $0.83$ on a four-combination split, and transfers at only $0.57$ to $0.59$. Four calibration rules on the saved scores leave held-out per-sample F1 at or below $0.678$. Finally, mosaics of $112$ px sub-crops from illumination-corrected pure-culture frames, $300$ images for each of the $26$ combinations of the five species with pure cultures in training and used only for training, cover six of the nine test combinations. The mosaic-trained decoder scores $0.52$ to $0.59$ AUROC on exactly those combinations against $0.60$ for the decoder trained on real images alone, and per-combination F1 agrees to within $0.03$. Coverage of the combination space is not the limit, because the frozen embedding of a mixture is not a mixture of its constituents' embeddings. All of these AUROC differences lie inside the intervals of Appendix~\ref{app:devprotocol}.

\section{Open World}
\label{app:openworld}

\subsection{LOOCV Protocol}
\label{app:loocv-algorithm}

Algorithm~\ref{alg:loocv} gives the leave-one-out cross-validation (LOOCV) harness of Section~\ref{subsec:openworld}. Features are extracted once and reused across folds, and each fold's inner loop takes under a second on one GPU, so the full six-fold sweep is quick and cheap to reproduce.

\begin{algorithm}[ht]
\caption{\phoebi{} LOOCV sweep for open-set rejection and discovery}
\label{alg:loocv}
\begin{algorithmic}[1]
\Require Train/val/test splits; tile config; $K$ species
\State Extract tile features on train/val/test once (cache-reuse across folds)
\For{$k^\star \in \mathcal{C}$}
    \State $\proto \gets \textsc{PureCultureInit}(\text{train}; \mathcal{C}\setminus\{k^\star\})$
    \State $\{\theta^A_k\}, \theta^A_{\mathrm{unk}}, \{\theta^B_k\}, \theta^B_{\mathrm{unk}} \gets \textsc{Calibrate}(\text{val}; \proto)$
    \State $(r_A, m_B, k\text{NN}) \gets \textsc{Score}(\text{test}; \proto)$ \Comment{open-set scores}
    \State $\proto_{\mathrm{new}} \gets \textsc{SinkhornCluster}_{K=1}(\{z_t : \lVert r_t\rVert > \theta_{\mathrm{disc}}\})$ \Comment{discovery; the gate admits every tile (Section~\ref{subsec:openworld})}
    \State Record AUROC, AUPR, FPR$@95$TPR; cluster accuracy and known-class drift
\EndFor
\State \Return per-fold and mean$\pm$std of all metrics
\end{algorithmic}
\end{algorithm}

\subsection{Open-Set Rejection}

\textbf{Open-set scoring functions.}\phantomsection\label{app:osr-sweep} All five scores in Table~\ref{tab:osr-sweep} share the same frozen features and pure-culture prototypes, so only the scalar read-out differs. The global scores are not just worse on average but unstable across folds, with a standard deviation of $0.15$ to $0.18$ against $0.066$ for the $k$-NN score, and Figure~\ref{fig:residual-hist} shows where that variance comes from. Table~\ref{tab:loocv} breaks the $k$-NN score and the residual down by fold, with one fold for each held-out species.

\input{tables/tab_osr_sweep}

\input{tables/tab_loocv}

\textbf{Open-set variants.}\phantomsection\label{app:negative} We reproduced the open-set and discovery results of Section~\ref{subsec:openworld} exactly and then varied three choices. The unit was a CLS tile or a patch token. The score was the residual, the negative maximum similarity, energy or $k$-NN, or the residual or $k$-NN divided by the median over validation units of the nearest known class. The aggregation over units was the mean, the $90$th percentile or the maximum. Selection used an inner leave-one-species-out loop over validation images. The paper's configuration, $k$-NN on CLS tiles averaged over the image, is the best of the $36$ (inner AUROC $0.74$). Class normalisation gives $0.67$, $90$th-percentile aggregation $0.63$, the maximum $0.59$, patch tokens $0.49$, and every residual, energy or negative-maximum variant $0.41$ to $0.48$.

\subsection{Discovery}

\textbf{The discovery gate.} The residual gate $\theta_{\mathrm{disc}} = 0.15$ admits every tile on every fold, so the $K = 1$ prototype is the mean of all test tiles. Gates that admit the top $5\%$ of validation known-tile scores reach $0.15$ cluster accuracy with the class-normalised residual on CLS tiles and $0.00$ on patch tokens.

\textbf{Gradient-based discovery heads.}\phantomsection\label{app:ncd-negative} We tested three more elaborate comparators. The first is a multi-label adaptation of UNO~\citep{fini2021unified}, a $[D \to K_{\mathrm{novel}}]$ head trained with BCE on Sinkhorn-Knopp one-hot pseudo-labels, whose normalised weights serve as the proposed prototypes, with the centroids refined by $100$ Adam steps of cosine-similarity BCE. At $K_{\mathrm{novel}} = 1$ it reaches $0.187 \pm 0.092$ cluster accuracy (drift $-0.013 \pm 0.004$), well below plain Sinkhorn-Knopp at $K = 1$ ($0.502 \pm 0.106$, drift $-0.031 \pm 0.006$). At $K_{\mathrm{novel}} = 4$ both variants fall to $0.0$ on all six folds, because the four refined slots drift off the cell-feature manifold and the sparsemax projection routes around them. The second is Cr-KD-NCD~\citep{gu2023class}, applied to Sinkhorn-Knopp at $K = 1$ as a drift-reducing refinement through a per-image gated distillation term on the known-class weights. At $\lambda = 1.0$ it collapses recall to $0$, and at $\lambda = 10^{-3}$ it lowers cluster accuracy to $0.110 \pm 0.19$ (drift $-0.030$). The cause is structural. Sparsemax produces exact zeros, so a prototype that does not reduce reconstruction error is dropped by the projection, and Sinkhorn-Knopp at $K = 1$ is both the simplest primitive that respects this and the strongest. A sparsemax-aware distillation term and the full UNO scaffold, which over-clusters and then converts head weights to prototypes, are natural extensions for future work beyond the scope of this paper.

The third is SimGCD~\citep{wen2023simgcd}, adapted as a $(K{-}1) + K_{\mathrm{novel}}$-way head with $K_{\mathrm{novel}} = 1$. Its known slots start from the pure-culture prototypes and its novel slot from the Sinkhorn-Knopp centroid at $K = 1$, and it trains on frozen \textsc{DINOv2}-S/14 features for $30$ epochs with the SimGCD recipe. That recipe applies supervised BCE on labelled training tiles to the $K{-}1$ known logits and adds a Sinkhorn-balanced soft pseudo-label cross-entropy on test tiles ($\tau_{\mathrm{sk}} = \tau_{\mathrm{ce}} = 0.1$, $w_{\mathrm{labelled}} = 0.5$). We drop the SupCon term, which needs augmentation pairs through the backbone, and keep the rest as published. With zero epochs the head is its Sinkhorn-Knopp initialisation and reproduces the $0.502$ cluster accuracy exactly. After five or more epochs the novel slot drifts away from the held-out species and cluster accuracy falls to $0.000$ on all six folds, identically for five and ten epochs at learning rates of $10^{-3}$ and $10^{-2}$. The balanced pseudo-label objective forces a uniform marginal across the six head columns, so the novel column follows whichever sub-population the balancing assigns to it rather than the held-out species, and with a single novel column nothing absorbs that noise. SimGCD's training loop suits the regime of many novel categories. When exactly one species is unknown, as in LOOCV, training is at best a no-op and at worst destabilises the initialisation, which is why plain Sinkhorn-Knopp at $K = 1$ is the primitive in Table~\ref{tab:discovery}.

%% file: tables/tab_dataset_card.tex
\begin{table}[h]
\centering
\small
\caption{\phoebi{} dataset card. A full datasheet~\citep{gebru2021datasheets} is in Appendix~\ref{app:datasheet}.}
\label{tab:dataset-card}
\begin{tabular}{>{\raggedright\arraybackslash}p{0.30\linewidth} >{\raggedright\arraybackslash}p{0.62\linewidth}}
\toprule
\textbf{Field} & \textbf{Value} \\
\midrule
Dataset name & \phoebi{}: Phase-contrast Optical bEnchmark for Bacterial Identification \\
Version & v1.0 (this paper) \\
Number of species ($K$) & $6$ \\
Number of combinations & $40$ ($6$ singles, $12$ pairs, $15$ triples, $6$ quadruples, $1$ six-species) \\
Number of images & $\sim\!120{,}000$ across the $40$ combinations \\
Image dimensions & $1024 \times 1024 \times 3$ (RGB) \\
Modality & Phase-contrast microscopy at $1000\times$ total magnification (100$\times$ oil-immersion objective) \\
Microscope make / model & Fisherbrand{\texttrademark} Advanced Research Grade Upright Phase Contrast Microscope (Trinocular model); CAT 03-000-106 \\
Camera make / model & Fisherbrand Microscope Camera (WiFi colour camera); CAT 03000045; CMOS sensor \\
Magnification / NA & 100$\times$ oil-immersion objective ($1000\times$ total) / NA $1.25$ (oil) \\
Illumination & K\"{o}hler-illuminated LED condenser, divide-by-Gaussian corrected \\
Acquisition sessions & Multiple per combination \\
Operators & $1$ \\
Sample preparation & Nutrient broth ($8\,\mathrm{g\,L^{-1}}$ in deionised water), autoclaved ($121\,^\circ\mathrm{C}$, $15\,\mathrm{min}$); inoculated from glycerol stock; incubated at $30\,^\circ\mathrm{C}$, $250\,\mathrm{rpm}$, $72$--$120\,\mathrm{h}$; growth monitored by colour change; imaged at characteristic growth stage \\
Species sourcing & ATCC 23857, DSM 13307, ATCC 25232, ATCC 13048, ATCC 17061, ATCC 13525 \\
Biosafety level & BSL-1 \\
\midrule
Label space & Multi-label binary, $\{0,1\}^6$, presence/absence per species \\
Label source & Folder name parsing (combination tokens) auto-discovered by \texttt{tools/build\_splits.py} \\
Combinations not collected & \texttt{bt}-\texttt{fj}, \texttt{bs}-\texttt{bt}, \texttt{ka}-\texttt{pf} (three pairs); plus all five-species combinations \\
Splits provided & In-distribution $80/10/10$, assigned in acquisition order within each combination (the ``random'' split); leave-combinations-out (seed $1337$, $9$ held-out combinations); five development folds (Appendix~\ref{app:devprotocol}) \\
\midrule
License \& hosting & CC-BY-4.0; Hugging Face dataset (Appendix~\ref{app:datasheet}) \\
Code repository & Public repository (Appendix~\ref{app:datasheet}) \\
\bottomrule
\end{tabular}
\end{table}

%% file: tables/tab_coverage.tex
\begin{table}[ht]
\centering
\small
\caption{Collected and absent combinations at each order. Over half of the gaps at orders $2$ to $4$ pair \texttt{bs} with \texttt{bt}, the one same-genus pair, which the collection avoided mixing.}
\label{tab:coverage}
\begin{tabular}{ccl}
\toprule
\textbf{Order} & \textbf{Collected} & \textbf{Absent} \\
\midrule
$1$ & $6/6$ & -- \\
$2$ & $12/15$ & \texttt{bs\_bt}, \texttt{bt\_fj}, \texttt{ka\_pf} \\
$3$ & $15/20$ & \texttt{bs\_bt\_fj}, \texttt{bs\_bt\_ka}, \texttt{bs\_bt\_pf}, \texttt{bt\_fj\_mx}, \texttt{bt\_ka\_pf} \\
$4$ & $6/15$ & $9$, of which $5$ contain \texttt{bs} and \texttt{bt} \\
$5$ & $0/6$ & all \\
$6$ & $1/1$ & -- \\
\bottomrule
\end{tabular}
\end{table}

%% file: tables/tab_label_noise.tex
\begin{table}[ht]
\centering
\small
\caption{Per-field label noise under \methodB{} on the clean test split. Each bound is the false-negative rate (FNR) in mixtures minus that in pure cultures, with the worst species in bold.}
\label{tab:label-noise}
\begin{tabular}{lccc}
\toprule
\textbf{Species} & \textbf{FNR, pure culture} ($n{=}300$) & \textbf{FNR, mixtures} & \textbf{Upper bound} \\
\midrule
\texttt{bs} & $0.0033$ & $0.0556$ ($n{=}4800$) & $0.0523$ \\
\texttt{bt} & $0.0200$ & $0.0450$ ($n{=}3600$) & $0.0250$ \\
\texttt{fj} & $0.0000$ & $0.0519$ ($n{=}5400$) & $0.0519$ \\
\texttt{ka} & $0.0000$ & $0.0676$ ($n{=}5400$) & $\mathbf{0.0676}$ \\
\texttt{mx} & $0.0000$ & $0.0448$ ($n{=}5400$) & $0.0448$ \\
\texttt{pf} & $0.0100$ & $0.0604$ ($n{=}5100$) & $0.0504$ \\
\midrule
mean & & & $0.0487$ \\
\bottomrule
\end{tabular}
\end{table}

%% file: tables/tab_grayscale.tex
\begin{table}[ht]
\centering
\small
\caption{Pure-culture test images scored in RGB and in grayscale by the frozen \methodB{}. \emph{Drop} is the fall in mean similarity to the species' own prototype, and \emph{margin} is own-prototype similarity minus similarity to the best other prototype.}
\label{tab:grayscale}
\resizebox{\linewidth}{!}{%
\begin{tabular}{lcccccccc}
\toprule
& \multicolumn{2}{c}{\textbf{Own-prototype similarity}} & & \multicolumn{2}{c}{\textbf{Pass rate at threshold}} & \multicolumn{2}{c}{\textbf{Margin}} & \\
\cmidrule{2-3} \cmidrule{5-6} \cmidrule{7-8}
\textbf{Species} & RGB & grayscale & \textbf{Drop} & RGB & grayscale & RGB & grayscale & $\Delta$\textbf{margin} \\
\midrule
\texttt{bs} & $0.732$ & $0.664$ & $0.068$ & $1.00$ & $0.18$ & $+0.006$ & $+0.010$ & $+0.005$ \\
\texttt{bt} & $0.705$ & $0.649$ & $0.056$ & $0.82$ & $0.10$ & $-0.024$ & $-0.007$ & $+0.017$ \\
\texttt{fj} & $0.727$ & $0.660$ & $0.066$ & $1.00$ & $0.45$ & $-0.011$ & $+0.003$ & $+0.014$ \\
\texttt{ka} & $0.917$ & $0.774$ & $0.144$ & $1.00$ & $1.00$ & $+0.127$ & $+0.142$ & $+0.015$ \\
\texttt{mx} & $0.777$ & $0.690$ & $0.087$ & $1.00$ & $1.00$ & $+0.042$ & $+0.042$ & $+0.000$ \\
\texttt{pf} & $0.748$ & $0.687$ & $0.061$ & $1.00$ & $0.28$ & $-0.008$ & $+0.000$ & $+0.008$ \\
\midrule
mean & & & $0.080 \pm 0.033$ & & & & & $+0.010$ \\
\bottomrule
\end{tabular}%
}
\end{table}

%% file: tables/tab_ablations.tex
\begin{table}[t]
\centering
\small
\caption{Pipeline ablations on the random test split, with the default variant marked in each block. The illumination rows are the matched arms of Appendix~\ref{app:acquisition}. The projection and tile-size rows come from separate runs with re-extracted features, so their default rows differ from Table~\ref{tab:headline} by up to half a point. Bold marks the best variant within each block.}
\label{tab:ablations}
\setlength{\tabcolsep}{5pt}   
\begin{tabular}{llcc}
\toprule
\textbf{Ablation} & \textbf{Variant} & \textbf{\methodA{} F1} $\uparrow$ & \textbf{\methodB{} F1} $\uparrow$ \\
\midrule
\multirow{2}{*}{Projection (\methodA{})}
 & softmax & $\mathbf{0.6200}$ (sparsity $0.00$) & $-$ \\
 & sparsemax (\emph{default}) & $0.6099$ (sparsity $0.11$) & $-$ \\
\midrule
\multirow{2}{*}{Illumination (matched arms)}
 & none (trained without) & $\mathbf{0.6172}$ & $\mathbf{0.6125}$ \\
 & divide (\emph{default}) & $0.6086$ & $0.6095$ \\
\midrule
\multirow{4}{*}{Tile size $s$}
 & $168$ & $0.6079$ & $\mathbf{0.6264}$ \\
 & $224$ (\emph{default}) & $\mathbf{0.6082}$ & $0.6139$ \\
 & $336$ & $0.5887$ & $0.6045$ \\
 & $518$ & $0.5943$ & $0.6065$ \\
\bottomrule
\end{tabular}
\end{table}

%% file: tables/tab_corruptions.tex
\begin{table}[ht]
\centering
\small
\caption{Change in per-sample F1 from the clean condition at the worst of three severities, for the frozen decoders with the deployed front-end. The severities are a ramp $1 \pm s$ across the field with $s \in \{0.15, 0.30, 0.50\}$ for the condenser gradient, a radial falloff $1 - s\,r^2/2$ with $s \in \{0.2, 0.4, 0.6\}$ for vignetting, a global scale of $\{0.85, 0.70, 0.55\}$ for lamp intensity, gamma $\{1.3, 1.6, 2.0\}$ for the detector, a Gaussian blur of $\sigma \in \{1, 2, 3.5\}$ px for focus drift, additive Gaussian noise of $\sigma \in \{5, 12, 25\}$ grey levels for the sensor and JPEG quality $\{60, 40, 25\}$. The last row replaces the tile grid with one resized tile. Clean F1 on the subsample is $0.6089$ for \methodA{}, $0.6096$ for \methodB{} and $0.6731$ for \methodC{}.}
\label{tab:corruptions}
\resizebox{\linewidth}{!}{%
\begin{tabular}{llccc}
\toprule
\textbf{Corruption} & \textbf{Acquisition variable} & \methodA{} & \methodB{} & \methodC{} \\
\midrule
Condenser gradient & condenser / K\"{o}hler misalignment & $-0.0004$ & $+0.0005$ & $-0.0005$ \\
Vignetting & objective / condenser aperture & $-0.0025$ & $+0.0021$ & $+0.0012$ \\
Lamp intensity & lamp voltage, exposure time & $-0.0014$ & $-0.0090$ & $-0.0224$ \\
Detector gamma & camera response curve & $-0.0016$ & $-0.0076$ & $-0.0071$ \\
\midrule
Focus drift & focus & $-0.0120$ & $-0.0833$ & $-0.0595$ \\
Sensor noise & camera, gain setting & $-0.1247$ & $-0.3531$ & $-0.1251$ \\
JPEG re-encoding & capture software, archiving & $-0.0475$ & $-0.3343$ & $-0.1735$ \\
\midrule
Resize to one tile & no tiling & $-0.3021$ & $-0.2105$ & $-0.2065$ \\
\bottomrule
\end{tabular}%
}
\end{table}

%% file: tables/tab_intervals.tex
\begin{table}[t]
\centering
\small
\caption{Combination-level intervals for the \phoebi{} decoders, on the three test partitions pooled (left) and under the development protocol (right). Margins are over the constant all-present predictor, and a species counts as resolved when its per-species AUROC interval excludes $0.5$.}
\label{tab:intervals}
\resizebox{\linewidth}{!}{%
\begin{tabular}{l ccc l ccc}
\toprule
& \multicolumn{3}{c}{\textbf{Test partitions pooled}} & & \multicolumn{3}{c}{\textbf{Development protocol}} \\
\cmidrule{2-4} \cmidrule{6-8}
\textbf{Decoder} & F1 margin & macro-F1 margin & mean AUROC & resolved (test) & F1 margin & macro-F1 margin & mean AUROC \\
\midrule
\methodA{} & $+0.006$ $[-0.008, +0.020]$ & $+0.011$ $[-0.003, +0.027]$ & $0.51$ $[0.46, 0.56]$ & \texttt{ka}; \texttt{mx} inverted & $-0.012$ $[-0.057, +0.032]$ & $-0.012$ $[-0.053, +0.026]$ & $0.53$ $[0.45, 0.61]$ \\
\methodB{} & $\mathbf{+0.029}$ $[+0.004, +0.053]$ & $\mathbf{+0.052}$ $[+0.030, +0.073]$ & $\mathbf{0.63}$ $[0.55, 0.71]$ & \texttt{ka}, \texttt{fj} & $+0.007$ $[-0.015, +0.030]$ & $+0.022$ $[-0.002, +0.048]$ & $0.55$ $[0.48, 0.61]$ \\
\methodC{} & $-0.020$ $[-0.083, +0.041]$ & $-0.002$ $[-0.055, +0.048]$ & $0.63$ $[0.54, 0.72]$ & \texttt{ka}, \texttt{mx} & $-0.077$ $[-0.132, -0.023]$ & $-0.086$ $[-0.146, -0.026]$ & $0.44$ $[0.36, 0.54]$ \\
\bottomrule
\end{tabular}%
}
\end{table}

%% file: tables/tab_osr_sweep.tex
\begin{table}[t]
\centering
\small
\caption{Open-set scoring functions under LOOCV, as mean $\pm$ std over the six folds. All five read the same frozen features, and bold marks the best per column.}
\label{tab:osr-sweep}
\begin{tabular}{l ccc}
\toprule
\textbf{Score function} & \textbf{AUROC} $\uparrow$ & \textbf{AUPR} $\uparrow$ & \textbf{FPR$@95$TPR} $\downarrow$ \\
\midrule
residual norm (\methodA{}) & $0.444 \pm 0.182$ & $0.428 \pm 0.110$ & $0.901 \pm 0.054$ \\
$-$max cosine (\methodB{})  & $0.444 \pm 0.181$ & $0.426 \pm 0.107$ & $0.902 \pm 0.049$ \\
energy ($T{=}1.0$)          & $0.437 \pm 0.154$ & $0.422 \pm 0.103$ & $0.931 \pm 0.047$ \\
energy ($T{=}0.1$)          & $0.444 \pm 0.170$ & $0.425 \pm 0.106$ & $0.917 \pm 0.047$ \\
\textbf{KNN ($k{=}10$)}     & $\mathbf{0.701 \pm 0.066}$ & $\mathbf{0.592 \pm 0.107}$ & $\mathbf{0.750 \pm 0.166}$ \\
\bottomrule
\end{tabular}
\end{table}

%% file: tables/tab_loocv.tex
\begin{table}[t]
\centering
\small
\caption{Open-set scores per LOOCV fold, each fold named by its held-out species. The $k$-NN score is the cosine distance to the tenth nearest training tile~\citep{sun2022knn}, shown beside the AUROC of \methodA{}'s residual, and bold marks the winner in each fold.}
\label{tab:loocv}
\resizebox{\linewidth}{!}{%
\begin{tabular}{l ccc c}
\toprule
\textbf{Held-out} & \textbf{$k$-NN AUROC} $\uparrow$ & \textbf{$k$-NN AUPR} $\uparrow$ & \textbf{$k$-NN FPR@95TPR} $\downarrow$ & \textbf{\methodA{} residual AUROC} \\
\midrule
\texttt{bs} & $\mathbf{0.636}$ & $\mathbf{0.534}$ & $\mathbf{0.904}$ & $0.585$ \\
\texttt{bt} & $\mathbf{0.688}$ & $\mathbf{0.420}$ & $\mathbf{0.688}$ & $0.390$ \\
\texttt{fj} & $\mathbf{0.660}$ & $\mathbf{0.619}$ & $\mathbf{0.899}$ & $0.491$ \\
\texttt{ka} & $\mathbf{0.835}$ & $\mathbf{0.747}$ & $\mathbf{0.424}$ & $0.066$ \\
\texttt{mx} & $\mathbf{0.726}$ & $\mathbf{0.686}$ & $\mathbf{0.741}$ & $0.582$ \\
\texttt{pf} & $\mathbf{0.659}$ & $\mathbf{0.548}$ & $\mathbf{0.844}$ & $0.551$ \\
\midrule
\textbf{Mean $\pm$ std} & $\mathbf{0.701 \pm 0.066}$ & $\mathbf{0.592 \pm 0.107}$ & $\mathbf{0.750 \pm 0.166}$ & $0.444 \pm 0.182$ \\
\bottomrule
\end{tabular}%
}
\end{table}